%% file: ms.tex
\documentclass[final,5p,times,twocolumn]{elsarticle}

\usepackage[utf8]{inputenc}
\usepackage{amsfonts}

\usepackage[ruled,vlined]{algorithm2e}

\usepackage{url}
\usepackage{hyperref}

\newtheorem{definition}{Definition}

\newcommand{\ie}{{i.e.,} }
\usepackage{listings,xcolor}
\usepackage{amsmath}
\usepackage{amssymb}
\DeclareMathOperator*{\argmax}{arg\,max}

\usepackage{xcolor}

\usepackage{booktabs}
\usepackage{color, colortbl}
\definecolor{Gray}{gray}{0.9}
\newcolumntype{J}{>{\centering\arraybackslash}m{0.65cm}}
\newcolumntype{K}{>{\centering\arraybackslash}m{0.7cm}}

\newcolumntype{M}{>{\centering\arraybackslash}m{0.80cm}}
\newcolumntype{S}{>{\centering\arraybackslash}m{0.85cm}}

\newcolumntype{P}{>{\centering\arraybackslash}m{0.90cm}}
\newcolumntype{T}{>{\centering\arraybackslash}m{0.95cm}}

\newcolumntype{L}{>{\centering\arraybackslash}m{1.0cm}}
\newcolumntype{D}{>{\centering\arraybackslash}m{1.1cm}}

\usepackage{soul}

\begin{document}

\begin{frontmatter}

\title{Attention-Based Model and Deep Reinforcement Learning for Distribution of Event Processing Tasks}

\author[ceot]{A. Mazayev\corref{amazayev}}
\ead{amazayev@ualg.pt}
\author[UK]{F. Al-Tam}
\ead{faroq.al.tam@gmail.com}
\author[ceot,ualg]{N. Correia}
\ead{ncorreia@ualg.pt}

\cortext[amazayev]{Corresponding author}

%%% Meta about authors %%%%%
\address[ceot]{Centre of Electronics, Optoelectronics and Telecommunications (CEOT), University of Algarve, 8005-139 Faro, Portugal}
\address[UK]{Departement of Computer Science, Aberystwyth University, United Kingdom}
\address[ualg]{Faculty of Science and Technology, University of Algarve, 8005-139 Faro, Portugal}

\begin{abstract}

Event processing is the cornerstone of the dynamic and responsive Internet of Things (IoT). Recent approaches in this area are based on representational state transfer (REST) principles, which allow event processing tasks to be placed at any device that follows the same principles.
However, the tasks should be properly distributed among edge devices to ensure fair resources utilization and guarantee seamless execution. This article investigates the use of deep learning to fairly distribute the tasks.
An attention-based neural network model is proposed to generate efficient load balancing solutions under different scenarios. The proposed model is based on the Transformer and Pointer Network architectures, and is trained by an advantage actor-critic reinforcement learning algorithm. The model is designed to scale to the number of event processing tasks and the number of edge devices, with no need for hyperparameters re-tuning or even retraining.
Extensive experimental results show that the proposed model outperforms conventional heuristics in many key performance indicators.
The generic design and the obtained results show that the proposed model can potentially be applied to several other load balancing problem variations, which makes the proposal an attractive option to be used in real-world scenarios due to its scalability and efficiency.

\end{abstract}

\begin{keyword}
Web of Things (WoT), representational state transfer (REST) application programming interface (APIs), Edge Computing, Load Balancing, Resource Placement, Deep Reinforcement Leaning, Transformer Model, Pointer Networks, Actor Critic
\end{keyword}

\end{frontmatter}

\section{Introduction}

Event processing is a crucial element in dynamic and reactive Internet of Things (IoT) applications, as it allows to derive real-time (or near-real-time) conclusions from data.
An event processing task, usually mapped into a web request (e.g., HTTP), follows a simple \texttt{observe-evaluate-actuate} pattern.
As the name suggests, this pattern involves performing three tasks: observe one or many devices/sensors, evaluate the produced data and, if the user-defined condition is satisfied, notify one or many actuators.
With the recent emergence of the Web of Things (WoT) \cite{duquennoy2009web,wotMashups,wotArch,ocfSpec} concept, where every device has its own internet protocol (IP) address and every device resource is addressable by uniform resource identifier (URI), the \texttt{observe-evaluate-actuate} chain operates over a set of URI endpoints.

Event processing has relied primarily on cloud computing, which virtually has an unlimited amount of storage and computational resources.
Centralized processing, alongside with the virtualization of both storage and computational resources, offers flexibility and efficiency by scaling up and adapting the system to different situations.
% allowing to quickly scale and adapt the system to different situations. 
However, with the ever-increasing amount of data being produced and consumed directly at the network edge, centralized (cloud) computing ends up becoming a limitation because additional IoT application requirements can not be satisfied.
% centralized (cloud) computing paradigm becomes a limitation as it is not able to satisfy additional requirements of new IoT applications. 
These requirements are related with delay sensitivity, data volume, uplink costs, non-interruption of service in environments with intermittent connectivity, privacy, and security\cite{iee_edge}.

The need to satisfy the above-mentioned requirements led to the emergence of a distributed computing paradigm, called edge computing.
In this new paradigm, substantial computational and data storage resources are located near the mobile devices, sensors and actuators. 
Edge computing can be seen as a strategy to provide a uniform computation/storage environment all the way from the core data centers to physical locations near users and data.
However, unlike the cloud, edge computing has a more dynamic and distributed computing nature, which results in distributed and ad-hoc event processing chains. 

Despite the above-mentioned differences, edge and cloud will have a common background in the underlying technologies.
In an Internet-Draft called ``IoT Edge Challenges and Functions" \cite{irtf-t2trg-iot-edge-01} Internet Engineering Task Force (IETF) emphasizes that ``\textit{virtualization platforms enable the deployment of virtual edge computing functions, including IoT gateway software, on servers in the mobile network infrastructure, in edge or regional data centers}". 
At the moment of writing, there are several initiatives that are moving in this direction.
As an example, Edge Virtualization Engine (EVE) \cite{eve} aims to create an open edge computing engine that enables the development, orchestration and security of cloud-native and legacy applications on distributed edge computing nodes.
It also aims to provide support for containers and clusters (e.g., Docker, Kubernetes), virtual machines, unikernels and offer a flexible foundation for IoT edge deployments with choice of any hardware, application and cloud. 
Another example is the EdgeX Foundry \cite{edgefoundry} service platform that also aims to provide management, data persistence or streaming near the edge.
This project also provides a dedicated rule engine API for the creation of event processing chains.
MobiledgeX \cite{mobiledgex} is an edge-cloud platform that provides an environment for the developers to deploy software (e.g., as software containers) on the edge. 
All these initiatives seek to bring orchestration, virtualization, load balancing techniques and technologies from the cloud to the edge.
This means that in order to understand how event processing tasks will be handled and distributed across the edge, it is important to see how this is currently being handled by the cloud.

\subsection{Adequate Resource Placement and Load Balancing} \label{sec:proxy_balance}

\begin{figure}\centering
\includegraphics[width=0.5\textwidth,trim={0.65cm 0.6cm 0.6cm 0.6cm}, clip]{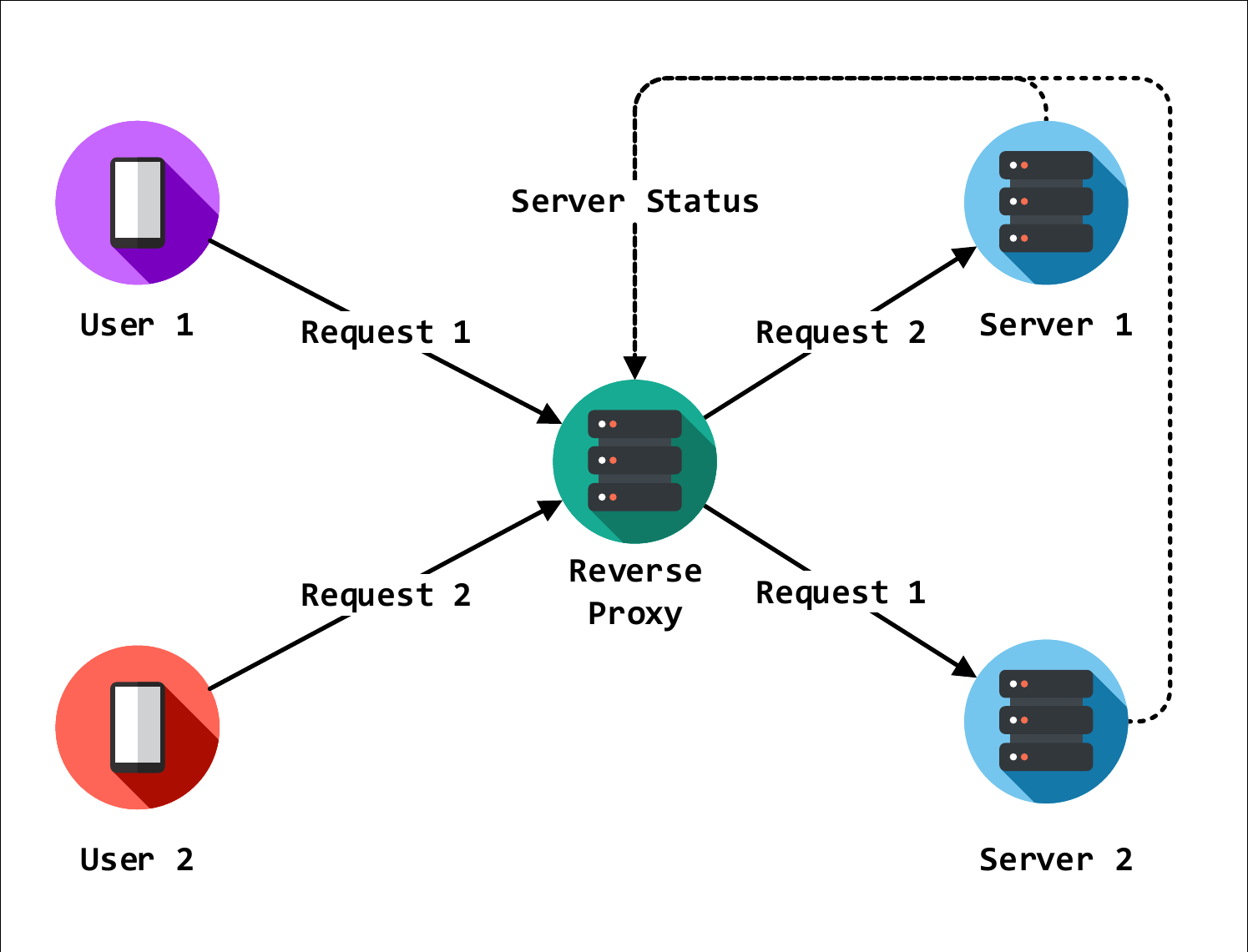}
\caption{An overview of reverse proxy.}
\label{fig:proxy}
\end{figure}

In typical application deployments, devices/servers (virtual or physical) are placed behind reverse proxies like NGNIX \cite{ngnix}, Traefik \cite{traefik} or Moleculer API Gateway \cite{moleculer}.
A reverse proxy, illustrated in \autoref{fig:proxy}, is a type of server that typically sits behind the firewall in a private network and distributes the incoming client requests among the available back-end servers.
Common features that reverse proxies provide include load balancing, web acceleration (e.g., caching, SSL encryption), security and anonymity. 
In case of load balancing there are several balancing strategies that can be used, each focusing on a specific aspect of Quality-of-Service (QoS).
NGNIX, for example, offers round-robin, least-connected, ip-hash \cite{ngnix_load}; Traefik at the moment only supports round-robin method\cite{traefik_load}; Moleculer API Gateway offers round-robin, random, CPU-usage based and sharding\cite{moleculer_load}. 
These load balancing strategies do not provide optimal solutions because these are prohibitively expensive to obtain in real-time.
Instead, these strategies trade the quality of solution for the response time, i.e., these strategies are fast but the solutions that they provide can be sub-optimal.
Moreover, static optimization strategies are not adequate because the environment is dynamic in terms of available servers and user requests.

% The arrival new user requests may be erratic, there might be bursts of requests during short periods of time followed by periods of slowdown.
The arrival of new user requests is usually irregular. That is, bursts of requests during short periods of time can be followed by periods of slowdown.
To address this, containerized environments have a container manager (e.g., Kubernetes) that provides autoscaling of the number of servers, ensuring the creation (or removal) of replicas according to the system load\cite{k8_autoscale}.
Similar dynamics are expected at the network edge, and projects like StarlingX\cite{starlingx} solve them by providing tools for management, orchestration and scaling of distributed edge-cloud computing devices.

%From the aforementioned discussion it is clear that load balancing and careful resource placement will play a crucial role in edge computing, including the placement of event processing tasks.
%New load balancing strategies must be scalable and adaptable to the dynamics of the systems: number of incoming requests, number of servers and resources available on the at that time.
%Generally speaking, load balancing boils down to the multiple knapsack problem, which is know to be NP-hard\cite{modenesi2008load}, therefore designing a handcrafted heuristic that quickly yields good results may be challenging, especially when the problem size increases.
%On the other hand, data-driven and machine learning (ML) approaches that offer a prospect of discovering new load balancing strategies, without the need of any hand-engineered reasoning, seem as a compelling choice and are envisioned as the right answer to an everincreasing complexity of the networks and the web\cite{feamster2017and}.

From the above discussion, it is clear that load balancing and resource planning strategies will play a crucial role in edge computing, together with strategies for the appropriate placement of event processing tasks.
Any strategy should be scalable and adaptable to the dynamics of the system: number of incoming requests, number of servers and resources available at that time.
In general, the problem of distributing event processing tasks boils down to the multiple knapsack problem, which is known to be NP-hard \cite{modenesi2008load}.
%%%%%%%%%%%%%%%%%%%%%%%%%%%%%%%%%%%%%% OLD %%%%%%%%%%%%%%%%%%%%%%%%%%%%%%%%%%%%%%%%%%%%%%%%%%
% Designing good handcrafted heuristics for such problems can be challenging, especially when the problem size increases.
% On the other hand, data-driven and machine learning (ML) approaches that, without much engineering, offer a prospect of discovering new distribution strategies seem to be a compelling choice.
% ML solutions are envisioned as the right answer to an ever increasing complexity of networks and the web \cite{feamster2017and}.
%%%%%%%%%%%%%%%%%%%%%%%%%%%%%%%%%%%%%% NEW %%%%%%%%%%%%%%%%%%%%%%%%%%%%%%%%%%%%%%%%%%%%%%%%%%
Developing handcrafted heuristics for such problems can be challenging, especially under dynamic environments and heterogeneous key performance indicators (KPIs).
On the other hand, machine learning (ML) solutions are envisioned as the right answer to an ever increasing complexity of networks and the web \cite{feamster2017and}.
ML approaches are able to learn new distribution strategies which makes them a compelling choice.

\subsection{Reinforcement Learning as a Solution Framework}

The most successful applications of ML methods (e.g. natural language processing (NLP), computer vision) fall under the umbrella of supervised learning (SL), a process that consists in learning to map a set of inputs to output labels.
However, applying supervised ML methods to placement problems, or any other combinatorial problem, is troublesome as it is almost always impossible to obtain the output labels.
Moreover, even if the labels are provided and a neural network model is trained in a supervised way, it usually has poor generalization to other problem instances \cite{bello2016neural}.

Reinforcement learning (RL), on the other hand, is an attractive approach to solve combinatorial problems because, unlike SL, it does not require labels.
RL can be seen as a trial and error process in which an agent interacts with an environment via a sequence of actions and, in return, receives feedback in the form of reward signals.
The goal of the agent is to learn an action selection strategy in a way that the total reward is maximized.
A continuous interaction with the environment allows the agent to explore several (millions of) possible environment configurations and to develop a strategy that ensures that the maximum reward is obtained in every possible situation, i.e., to generalize.

Recent attempts in integrating deep learning (DL) with RL showed breakthrough results in video games \cite{mnih2015human}, board games \cite{silver2016mastering, schrittwieser2020mastering}, robotics and many more areas \cite{arulkumaran2017deep}.
This new learning paradigm is widely known as deep reinforcement learning (DRL).
In this work we investigate how DRL can be used by reverse proxy servers to distribute event processing tasks among heterogeneous edge devices available for processing.

\subsection{Motivation and Contributions}

The idea of using DRL to solve load balancing and distribution problems has recently attracted many researchers.
Surveys show that there are several attempts, with different levels of success, to solve these kind of problems \cite{shakarami2020survey, lei2020deep}.
However, the vast majority of these attempts do not solve problems of variable sizes.
The problem is treated as a fixed-sized optimization problem where a neural network is trained to map problem instances to solutions.
For instance, in the distribution problem illustrated in \autoref{fig:proxy}, the number of incoming requests and the number of available servers are assumed to be fixed.
Then, during the development (design and training), the model is trained to learn to map the inputs to outputs.
This is followed by testing, where the network's performance is measured on solving problems of the same size that it was trained upon and, finally, reporting the obtained results.
However, despite the results reported in the above mentioned surveys, the majority of those models have little practical use in real life deployment because a simple change in input size would make those models invalid.
% These vast majority of proposed models are not capable to adapt to variable input size.
% Adapting to a different input size would imply retraining the model, a process that might take several hours.

We argue that the assumption of a fixed problem size is problematic and, usually, does not hold true in dynamic environments like IoT and edge.
The number of incoming requests and available nodes is dynamic and changes over time.
Therefore, the main limitation of the existing literature that involves DRL, load balancing and distribution problems is the inability of the proposed neural networks to adapt to the dynamics of the environment without retraining.
In other words, the problem is the scalability of the proposed neural networks.

This works focuses on developing a scalable DRL model that is able to solve problem instances of variable size without retraining.
The main contributions of this article are as follows:
\begin{enumerate}

\item A study is made on how event processing tasks can be distributed across edge devices. Three different distribution criteria are studied, each being formalized mathematically.

\item A new neural network architecture is proposed for the distribution of event processing tasks, while ensuring load balancing. 
The key feature of the proposed model is its scalability, which means that it does not require retraining every time the number of incoming requests or servers, available for processing, change.

\item A comparative study against the optimal solutions and several baseline heuristics is carried out.
Obtained results indicate that the proposed model can generate high quality solutions to problems up to five times larger that the ones it was trained upon.

\end{enumerate}

The remainder of this article is organized as follows.
In Section \ref{sec:relate_work}, existing attempts of using RL in load balancing problems are discussed. A review of neural combinatorial optimization is also carried out.
The event task distribution problem is mathematically formulated in Section \ref{sec:math_load_balancing}.
Section \ref{sec:framework} presents the proposed neural network architecture and is organized as follows:
$i$) Section \ref{sec:RL_basics} presets a brief RL and DNN background;
$ii$) Section \ref{sec:mdp_load_balancing} presents the Markov decision process (MDP) representation of the problem; 
$iii$) Section \ref{sec:rl_load_balancing} describes the detailed design of the proposed neural network architecture.
Section \ref{sec:perfomance} describes the training method alongside with the hyperparameters used to obtain the model and evaluates its performance. 
Finally, in Section \ref{sec:conclusions} conclusions are drawn.

\section{Related Work} \label{sec:relate_work}

RL has been applied in several areas of IoT and edge computing for different purposes, which range from the actuation control (e.g, greenhouse climate control) to resource control (e.g., minimization of communication delay, energy consumption, hardware resources) \cite{shakarami2020survey, lei2020deep}.
However, most of the literature uses deep neural network (DNN) architectures that are not able to scale or adapt to the dynamics of the environment, e.g., increase in the number of edge devices, base stations, sensors or number of requests.
That is, the action space is fixed, which limits the practical use of DNNs in realistic scenarios.
The following subsections provide an overview of the application of RL to IoT, edge computing and load balancing.
Then, ongoing research to tackle combinatorial optimization problems, using RL, is presented \cite{bengio2020machine, mazyavkina2020reinforcement}.

\subsection{Reinforcement Learning in Edge Computing}

In \cite{chiang2020deep} authors propose a DRL solution to minimize the total cotask completion time. A cotask is a task, generated by an IoT device, that can only be completed when all of its constituent sub-tasks are finished.
Authors use a DNN architecture with two heads, one that predicts the offloading location of the task and another for the prediction of the cotask completion time.
The obtained results show that the proposed model is able to outperform several baseline heuristics.
However, the proposed architecture requires retraining whenever the network or the input size changes.

In \cite{zhang2021new} an RL method is used to distribute the tasks, produced by different user devices, among a set of edge servers.
To deal with the combinatorial action space (any task can be placed at any available edge server), authors introduce a multi-agent algorithm where each Deep Q-Network (DQN) \cite{mnih2015human} agent makes an action that corresponds to the location where the task will be offloaded.
The DQN architecture, used in this work, requires retraining once the state or action space changes, i.e., once an additional edge server is connected to the system.

In \cite{xu2019deep} mobility load balancing in self-organizing networks is investigated.
The goal is to design a handover scheme that transfers a mobile user from its service cell to a neighbor cell that can handle the incoming traffic.
The authors use an off-policy deterministic policy gradient (OPDPG) method \cite{silver2014deterministic} and propose an asynchronous parallel learning framework to improve the training efficiency in a collaborative manner.
The obtained results show that the RL method is able to outperform several baseline heuristics.
% However, once again, this neural architecture only works for static environments, i.e., adding an additional cell implies retraining the network.
However, once again, this architecture is designed for fixed-size communication networks, i.e., any change in the number of cells requires retraining the network.

In \cite{zhang2018joint} the authors propose an RL agent to offload computationally intensive tasks, generated by user equipment (UE), to mobile edge computing (MEC) servers, while maximizing the mobile operator revenue and minimizing the energy consumption and time-delays.
In the investigated environment the authors considered a queuing model, states of energy harvesting batteries and down-link transmit power costs.
The authors conclude that their approach has better performance than the policy-gradient and Q-learning algorithms.
The main limitation of this work is the fact that the authors only studied a scenario with a fixed number of MEC servers, which is not always the case in reality. %Authors state that ``\textit{multi-UE case will be discussed in our future work}".
% The approach presented in this work means that an addition or removal of a MEC would require retraining the agent.
Moreover, any change in the number of MEC servers requires retraining the agent.

Several other works \cite{alam2019autonomic, chen2018optimized, zhang2019deep, yang2018deep, van2018deep} used DQN-based models in offloading and load balancing problems.
Despite the obtained results, all of these proposals have a common weakness that drastically limits their use in real world deployments. Any change in the state or action space sizes imply retraining the network from scratch.
Overall, DQN and its variants are more suitable for static environments with fixed state and action spaces but they have limited uses in dynamic environments

\subsection{Combinatorial Optimization with Reinforcement Learning} \label{sec:related_combo_rl}

One of the first attempts to solve combinatorial problems with neural networks was presented in \cite{vinyals2015pointer}.
The authors propose a neural architecture, called Pointer Networks (Ptr-Net), which is based on sequence-to-sequence (seq2seq) \cite{sutskever2014sequence} models with attention mechanisms \cite{bahdanau2014neural, luong2015effective} that are commonly used in NLP.
Ptr-Net is an encoder-decoder architecture where recurrent neural networks (RNNs), usually long short-term memory network (LSTM) \cite{hochreiter1997long} or gated recurrent unit (GRU) \cite{cho2014learning}, are used to process the input sequence at the encoder and to generate the output sequence at the decoder.
Although in this work the Ptr-Net was trained in a supervised way, authors showed that a single architecture with the same hyperparameters can be used to solve different combinatorial problems.
The fundamental breakthrough of Ptr-Net is its ability to deal with variable output space without the need for retraining the network.
Moreover, the authors showed that the Ptr-Net trained on combinatorial problems of small size is capable to generalize to much larger problems without considerable degradation in the quality of generated solutions.

%%%%

In \cite{bello2016neural} the authors used Ptr-Net in combination with RL to solve combinatorial problems, such as the traveling salesman problem (TSP) and the knapsack problem. 
Authors use RL methods because it is difficult to have access to labels (associated with optimal solutions), as combinatorial problems are usually NP-Hard.
% Authors use RL methods because it is difficult to have access to optimal labels of combinatorial problems, which are usually NP-Hard.
The obtained results show that the developed RL agent is capable of finding near-optimal solutions for TSP with up to 100 nodes and for up to 200 items in knapsack problem.
In this work the authors also made an empirical analysis between supervised training and RL and reached the conclusion that Ptr-Net networks trained in a supervised way have poor generalization when compared to a Ptr-Net-based RL agent that explores different solutions and observes their corresponding rewards.

In \cite{nazari2018reinforcement} the authors proposed a model, inspired on Ptr-Nets and seq2seq, to solve vehicle routing problem (VRP).
Authors removed LSTM from the encoder as they argue that LSTM, and other RNNs, are only necessary when dealing with sequential data.
After training, the proposed model was able to generalize well and generate high-quality solutions for all problems sampled from the same distribution that was used during training.
Moreover, authors tested the model's ability to handle variable problem sizes.
They conclude that their model performs well when training and testing problem sizes are close to each other.
However, they report a degradation in performance when the sizes of the testing instances are substantially different from the ones used for training.

% In \cite{kool2018attention} extended the work presented in \cite{nazari2018reinforcement} and proposed a model capable of solving several variations of VRP and TSP.
% Authors replaced LSTM networks with the Transformer\cite{vaswani2017attention} and the obtained results show that the proposed model achieves (near) optimal solutions for problems of size 20, 50 and 100.
The work in \cite{nazari2018reinforcement} is extended by \cite{kool2018attention}, where a model capable of solving several variations of VRP and TSP is proposed. The authors replace LSTM networks by the Transformer\cite{vaswani2017attention} and results show that the proposed model is able to achieve (near) optimal solutions for problems of size 20, 50 and 100 nodes.

%%%%%%%%%%%%%%%%%%%%%%%%%%%%%%%%%%%% OLD %%%%%%%%%%%%%%%%%%%%%%%%%%%%%%%%%%%%
%\hl{Overall, the above mentioned models were designed to tackle ``classical" combinatorial optimization problems. However, they did not explore their generalizations involving multiple elements, i.e., multiple knapsack problem, multiple TSP or multiple VRP. To be able to solve such problems, the proposed models would have to be redesigned and adapted.}
%In \cite{bello2016neural,nazari2018reinforcement,kool2018attention, vaswani2017attention}, \hl{this type of problems were either ignored or left for future work.}

\section{The Problem of Event Processing} \label{sec:math_load_balancing}

Prior to presenting the proposed DRL solution, it is necessary to clearly define the problem.
This section briefly describes the technological background of event processing, which is then followed by its mathematical formalization.

\subsection{Rule Synchronization Mechanism}

There are multiple Internet-Drafts that attempt to standardize event processing in WoT, many of them have been reviewed and analyzed in our previous work \cite{mazayev2019distributed}.
In this article the focus will be on the most recent proposal, called \texttt{Rule}, which was also discussed in\cite{mazayev2019distributed}.

\begin{figure}\centering
\includegraphics[width=0.5\textwidth,trim={0.6cm 0.5cm 0.5cm 0.5cm}, clip]{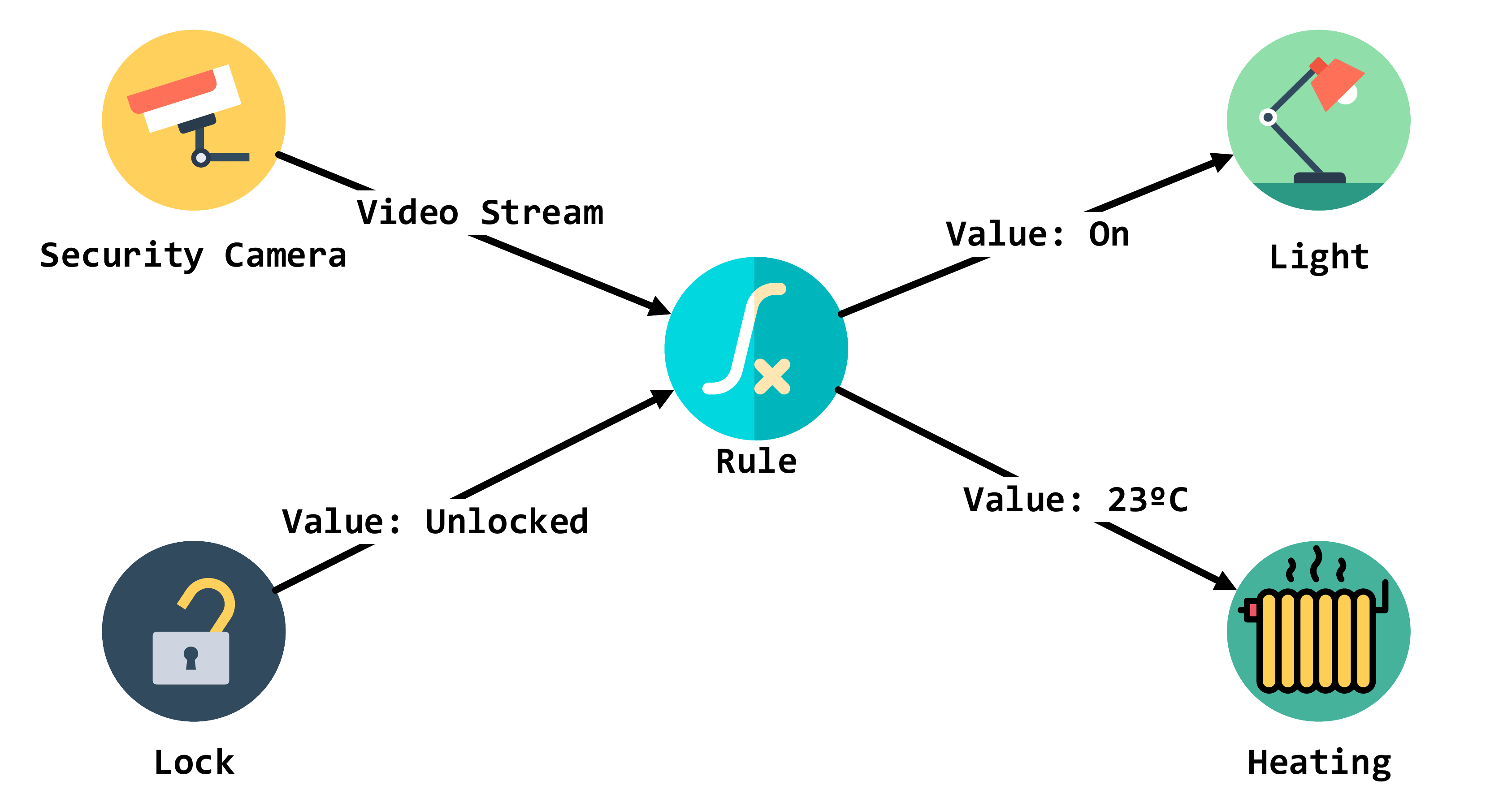}
\caption{An example of a \texttt{Rule} with two inputs and two outputs.}
\label{fig:rule}
\end{figure}

A \texttt{Rule}, illustrated in \autoref{fig:rule}, is a regular web resource, based on RESTful principles, that allows to create complex many-to-many event processing chains.
Multiple sources can be observed (e.g., sensors) and, after condition evaluation, multiple destination endpoints can be updated (e.g., actuators).
Being a regular web resource means that \texttt{Rules} can be placed at any device that supports a \texttt{Rule} API and is available for processing.
The \texttt{Rule} is a continuously running task that is constantly observing resources and processing the data being sent.
To stop it, and to release the resources that it is consuming, the user must send an appropriate request (e.g., HTTP DELETE).

End-users are not expected to create \texttt{Rules} by hand.
Applications are expected to be available (e.g., \texttt{Rules} store) where users can find pre-defined \texttt{Rules}, together with related meta-data (e.g., CPU, RAM and storage requirements), i.e., \texttt{Rule} profiles.
The inputs and outputs of a pre-configured \texttt{Rule} will have to be defined according to user's need.

Running \texttt{Rules} at the edge, where there will be physical, virtual, and containerized nodes with different computational capabilities, which means that these tasks must be carefully distributed.

\begin{definition}[Rule Distribution (RD) Problem]\label{def:RD_problem} Given a set of \texttt{Rule} profiles, decide for the best \texttt{Rule} distribution across devices while taking into account randomly arriving \texttt{Rule}  requests, random number of available edge devices (each having its random amount of available CPU, RAM and storage resources) and QoS requirements.
\end{definition}

\subsection{Assumptions and Notation}
To have a clear understanding of how to solve the RD problem optimally, assumptions and notation have to be clarified. This is followed by the mathematical formalization of the RD problem considering three variants. Such mathematical formalization is important not only to clearly define the problem, but also to have a reference when evaluating the proposed RL-based agent.

\begin{definition}[Node]\label{p:Node} A virtual, containerized or physical device exposing a \texttt{Rule} API, being capable of processing a specific set of \texttt{Rules}. The set of nodes able to host \texttt{Rule} $x$ is denoted by $\mathcal{N}^x$, while the overall set of heterogeneous nodes is denoted by $\mathcal{N}$.
\end{definition}

\begin{definition}[Computational Resource]\label{p:Compute} Hardware resource (e.g., CPU, RAM, memory) available at nodes. The amount/capacity of $n$'s computational resources, $n \in \mathcal{N}$, is  given by vector $\textbf{c}^n=[c_{n,m}]$, where $m \in \{1,...,M\}$ and $M$ is the total number of computational resources considered.
\end{definition}

\begin{definition}[Rule]\label{p:Rule} Web resource able to evaluate an expression or performing complex processing over certain inputs. Assuming $\mathcal{X}=\{x_1,x_2,...,x_{|\mathcal{X}|}\}$ to be a set of \texttt{Rules}, each $x_i$ has a demand vector $\textbf{d}^{x_i}=[d_{x_i,m}]$, $m \in \{1,...,M\}$,  containing computational resource requirements.
\end{definition}

\begin{definition}[Critical Resource]\label{p:DominantResource} Given a set of rules $\mathcal{X}'\subset \mathcal{X}$ under execution at node $n \in \mathcal{N}$, $n$'s critical resource is given by $\Omega^n=\argmax_{m \in \{1,...,M\}} \{\frac{\sum_{x' \in \mathcal{X}'}{d_{x',m}}}{c_{n,m}}\}$.
\end{definition}

In the following section three variants of the RD problem are formalized. Each variant tries to optimize specific performance indicators: number of allocated \texttt{Rules}, critical resource and/or number of active nodes. The variables that may be involved are the following:

\begin{tabular}{p{1.1cm} p{6.8cm}}
$\omega_{x}$ & One if \texttt{Rule} $x \in \mathcal{X}$ is to be executed, which requires finding a node to place it; zero otherwise.\\
$\beta^n_{x}$ & One if \texttt{Rule} $x \in \mathcal{X}$ is to be placed at node $n \in \mathcal{N}^{x}$; zero otherwise.\\
$\Omega^n$ & Critical resource at node $n \in \mathcal{N}$.\\
$\Omega^{\text{MAX}}$ & Most critical resource among all nodes at the network.\\
$\Theta^n$ & One if node $n \in \mathcal{N}$ is in use; zero otherwise.
\end{tabular}

\subsection{Mathematical Formalization} \label{opt:probs}

\subsubsection{Greedy Optimizer} \label{cplex:greedy}
The performance indicator to be optimized is the number of \texttt{Rules} placed/distributed across nodes, which should be maximized. This is expressed as follows.

-- Objective Function:

\begin{align}
\text{Maximize}\ \sum_{\{x \in \mathcal{X}\}} \omega_{x}
\end{align}

-- Placement of Rules:

\begin{align} \label{eq:placement}
\sum_{\{n \in \mathcal{N}^{x}\}} \beta^n_{x} = \omega_{x}, \forall x \in \mathcal{X}
\end{align}

These constraints ensure that there is a single location for each \texttt{Rule} that is to be executed.

-- Computational Resource Limitation:

\begin{align} \label{eq:comp_limitation}
\sum_{\{x \in \mathcal{X}\}} d_{x,m} \times \beta^n_{x}\leq c_{n,m}, \forall n \in \mathcal{N}, \forall m \in \{1,...,M\}
\end{align}
where $d_{x,m}$ is known information. These constraints avoid exceeding the available computational resources of nodes. 

-- Non-negativity assignment to variables:

\begin{align}
\omega_{x}, \beta^n_{x} \in \{0,1\}.
\end{align}

\subsubsection{Critical-Aware Greedy Optimizer} \label{cplex:crit_aware}
The main performance indicator to be optimized is the number of \texttt{Rules} that are placed/distributed across nodes, which should be maximized while ensuring that load is fairly distributed (most critical resource as a reference). Thus, fair distribution is the secondary performance indicator to be optimized. This kind of optimization is important in cooperative scenarios when edge devices are provided by different owners that trade their computational resources for some reward (e.g., financial benefits). Service providers hold reverse proxy servers, which exposes the \texttt{Rule} API to the end-users. The computational resources are provided by the device owners.

-- Objective Function:

\begin{align}
\text{Maximize}\ \sum_{\{x \in \mathcal{X}\}} \omega_{x} + \Omega^{\text{MAX}}
\end{align}

This objective function ensures that the largest number of \texttt{Rules} is placed/distributed and, at the same time, that it is done in a fair way, so that \texttt{Rules} have the best (and similar) working conditions.

-- Placement of Rules: Expression (\ref{eq:placement}).

-- Computational Resource Limitation: Expression (\ref{eq:comp_limitation}).

-- Obtaining Critical Resource:

\begin{align}
\Omega^n \leq c_{n,m} - \sum_{\{x \in \mathcal{X}\}} d_{x,m} \times \beta^n_{x}, \forall n \in \mathcal{N}, \forall m \in \{1,...,M\}
\end{align}
where $d_{x,m}$ is known information. These constraints finds the critical resource at each node. Finding the most critical resource among all nodes requires:

\begin{align} \label{eq:most_critical}
\Omega^{\text{MAX}} \leq \Omega^n, \forall n \in \mathcal{N}
\end{align}

-- Non-negativity assignment to variables:

\begin{align}
\omega_{x}, \beta^n_{x} \in \{0,1\}; 0 \leq \Omega^n, \Omega^{\text{MAX}} \leq 1.
\end{align}

\subsubsection{Cost-Aware Greedy Optimizer} \label{cplex:node_aware}
The main performance indicator to be optimized is the number of \texttt{Rules} that are placed/distributed across nodes, which should be maximized while ensuring that the number of nodes in use (hardware cost) is minimized. Thus, the cost is the secondary performance indicator to be optimized. This optimization is important when the \texttt{Rule} API provider is responsible for both the reverse proxy and nodes, but is using rented hardware to offer its services. This is a very common scenario in cloud computing, so rental cost minimization ends up being also a relevant performance indicator.

-- Objective Function:
\begin{align}
\text{Maximize}\ \sum_{\{x \in \mathcal{X}\}} \omega_{x} - \frac{\sum_{\{n \in \mathcal{N}\}}\Theta^n}{|\mathcal{N}|}
\end{align}

This goal ensures that the largest number of \texttt{Rules} is placed/distributed using the lowest number of nodes, for cost minimization.

-- Placement of Rules: Expression (\ref{eq:placement}).

-- Computational Resource Limitation: Expression (\ref{eq:comp_limitation}).

-- Obtaining nodes in use:

\begin{align}
\Theta^n \geq \beta^n_{x}, \forall n \in \mathcal{N}, \forall x \in \mathcal{X}
\end{align}

\begin{align}\label{eq:exec_avail_resources}
\sum_{x \in \mathcal{X}} d_{x,m} \times \beta^n_{x}\leq c_{n,m}, \forall n \in \mathcal{N}, \forall m \in \{1,...,M\}
\end{align}
Constraints (\ref{eq:exec_avail_resources}) avoid exceeding the available resources at nodes. 

-- Non-negativity assignment to variables:

\begin{align}
\omega_{x}, \beta^n_{x}, \Theta^n \in \{0,1\}.
\end{align}

The above-mentioned optimization problems provide the optimal solution for each problem instance.
However, solvers (e.g., CPLEX\cite{cplex_cite}, Gurobi\cite{gurobi_cite}) that rely on these formalizations to compute optimal solutions can only find results, in a reasonable amount of time, if the problem instance is very small.
For this reason an RL-based solution is proposed.
RL methods have the ability to understand the system dynamics and, after trained, make appropriate placement decisions in real-time.

\section{Proposed DRL Framework} \label{sec:framework}

Prior to presenting the proposed DRL solution designed to tackle the RD problem, formalized in \autoref{sec:math_load_balancing}, this section first presents a brief RL and DNN background, which is then followed by the MDP formulation of RD problem.

\subsection{Deep Reinforcement Learning Background}\label{sec:RL_basics}
\subsubsection{Reinforcement Learning} 

In RL, sequential decision problems can be modeled using Markov decision processes (MDPs). 
An MDP can be represented by the tuple $(\mathcal{S}, \mathcal{A}, p, r)$, where $\mathcal{S}$ is the state space and $\mathcal{A}$ is the action space. 
The $p:\mathcal{S}\times\mathcal{A}\times\mathcal{S}\rightarrow [0, 1]$ is the probabilistic transition model (matrix), where $p(s^{\prime}|s,a)$ is the probability of arriving at state $s^{\prime}$ from $s$ after taking action $a$, of which $s^{\prime}, s \in \mathcal{S}$ and $a\in \mathcal{A}$; $r:\mathcal{S} \times \mathcal{A} \times \mathcal{S} \rightarrow \mathbb{R}$ is the reward function.

An agent learns by interacting with the environment. At time step $t$, the agent observes a state $s_t$ and takes an action $a_t$. The environment returns a reward signal $r(s_t, a_t)$ to the agent and makes a transition to a next state $s_{t+1}$. The objective of the agent is to maximize the collected reward (outcome):

% \begin{equation}\label{equ:outcome}
% G_t = \mathbb{E}\left[\sum\limits_{k = 0}^{\infty} \gamma^k r(s_{t+k}, a_{t + k})| s_0 = s_t\right]
% \end{equation}

\begin{equation}\label{equ:outcome}
G_t = \left[\sum\limits_{k = 0}^{\infty} \gamma^k r(s_{t+k}, a_{t + k})| s_0 = s_t\right]
\end{equation}

where $\gamma$ is a discount factor used to balance the importance of immediate and long-term rewards.

For the agent to decide which action to take at a given state, it needs to follow a policy.
Therefore, the objective of training an agent is to find a policy $\pi$ that maximizes $G_t$.
A policy is optimal $\pi^*$ if following it produces the maximum outcome.
In addition, a policy $\pi$ can be deterministic \ie $\pi(s) = a$, or stochastic \ie a probability distribution over the action space $\mathcal{A}$.

There are two ways to learn $\pi$: value-based and policy-gradient. This work focuses only on the latter.

In value-based learning, $\pi$ is calculated from a state-value function $V_{\pi}(s)$ which measures how good it is to be at a given state $s$:

\begin{equation}\label{equ:value_function}
V_{\pi}(s) = \mathbb{E}\left[G_t | s_t = s \right]    
\end{equation}

%\subsection{Policy gradient} \label{sec:pg_ac} 
In policy gradient learning, $\pi$ can be modeled by a parametric function where the objective is to learn a set of parameters $\theta$ that maximizes an objective function $J$. 
Assume an episodic setting with episode length $T$ and let $\tau$ be a sequence of transitions (trajectory) $\tau = s_0, a_0, \dots, s_{T - 1}, a_{T - 1}$. Following the probability general product rule, $\pi$ can be defined as:

\begin{equation}\label{equ:policy}
\pi(\tau;\theta) = p(s_0) \prod\limits_{t = 0}^{T - 1} \pi(a_t|s_t; \theta) p(s_{t+1}|s_t, a_t)
\end{equation}

where $p$ is, again, the transition probability density function and $\theta$ is the set of parameters. 

From Expression (\ref{equ:policy}), it is possible to see that the transition from $s$ to $s_{t+1}$ depends on the action taken, $a_t$, and the probability of reaching $s_{t+1}$ from $s_t$ when taking $a_t$. 
Furthermore, following the original assumption about $\pi^*$: \emph{$\pi^*$ will generate the optimal trajectory $\tau^*$ that collects the maximum accumulated reward}, the objective function $J$ can be formulated as:

\begin{equation}
J(\theta) = \mathbb{E}_{\tau \sim \pi_{\theta}} [r(\tau)] = \int \pi(\tau;\theta) r(\tau),
\end{equation}
and the optimal set of parameters can be found using:
    
\begin{equation}\label{equ:theta}
\theta^* = \arg\max J(\theta)
\end{equation} 

%Please note that we are maximizing in \autoref{equ:theta} since we are seeking the maximum accumulated outcome from the episode. To find $\theta$, we need first to simplify the product in \autoref{equ:policy} and then differentiate with respect to $\theta$. We can do both using the logarithmic differentiation:

To find $\theta$, the product in Expression (\ref{equ:policy}) has to be simplified and differentiated with respect to $\theta$.
Both can be done using the logarithmic differentiation:

\begin{align}\label{equ:exp_pg}
& \nabla_\theta J(\theta) = \mathbb{E}_{\tau \sim \pi_{\theta}}\left[\nabla \log(\pi(\tau;\theta))  r(\tau)\right]
\end{align}

Moreover, since $r$ does not depend on $\theta$, we have:

\begin{equation}\label{equ:reinforce_eq}
    \nabla_\theta J(\theta) = \mathbb{E}_{\tau \sim \pi_{\theta}} \left[\sum\limits_{t = 0}^{T - 1} \nabla \log(\pi(a_t| s_t; \theta))  G_t\right]
\end{equation}

    % \begin{equation}
    %   \nabla_\theta J(\theta) = \frac{1}{M} \sum\limits_{i = 1}^M \left( \sum\limits_{t = 1}^N \nabla_\theta \log\pi(a_{i,t}|s_{i,t}; \theta)\right) \sum\limits_{t = 1}^T r(s_{i, t}, a_{i,t})
    % \end{equation}

% Following the gradient ascent rule (since we are maximizing) we have the update step:
Following the gradient ascent rule, the update step is:

\begin{equation}\label{equ:update_theta}
\theta_{k+1} = \theta_{k} + \alpha \nabla_{\theta}J(\theta_k)
\end{equation}

where $\alpha$ is the learning rate.

The steps of: $i)$ sampling a trajectory $\tau \sim \pi_{\theta}$; $ii)$ differentiating with respect to $\theta$ (Expression (\ref{equ:reinforce_eq})); $iii)$ updating $\theta$ (Expression (\ref{equ:update_theta})); are known as the REINFORCE algorithm \cite{ISI:A1992HV72300002, Sutton1998}.
This method, however, can suffer from high gradient variance due to trajectory sampling.
A common way to reduce the variance is by subtracting a baseline value from the reward term in Expression (\ref{equ:exp_pg}).
When the baseline is used, Expression (\ref{equ:reinforce_eq}) can be rewritten as:

\begin{equation}\label{equ:reinforce_eq_A}
\nabla_\theta J(\theta) = \mathbb{E}_{\tau \sim \pi_{\theta}} \left[\sum\limits_{t = 0}^{T - 1} \nabla \log(\pi(a_t| s_t; \theta))  A(s_t, a_t)\right], 
\end{equation}

where $A(s_t, a_t)$ is known as the advantage function \cite{10201}:
 
\begin{equation}\label{equ:advantage}
A(s_t, a_t) \approx r(s_t, a_t) + \gamma V_{\pi_{\theta}}(s_{t + 1}) - V_{\pi_{\theta}}(s_t)
\end{equation}

and $V_{\pi_{\theta}}(s_t)$ is the actual baseline value (Expression \eqref{equ:value_function}).

The method using the Expression \eqref{equ:reinforce_eq_A} is called advantage actor critic (A2C) \cite{pmlr-v48-mniha16}.
As the name suggests, this algorithm relies on an actor to learn the $\pi_\theta$ and on the critic to learn the baseline values.
An efficient way to model both the actor and the critic is by using DNNs.
In this case, their losses can be defined as follows:

\begin{align}
\text{actor loss: } - \sum\limits_{t = 0}^{T  -1 } \log(\pi(a_t|s_t;\theta)) A(s_t, a_t)
\end{align}

\begin{align} \label{eq:entropy}
\text{entropy: } -\sum\limits_{t = 0}^{T - 1} \pi(a_t|s_t) \log(\pi(a_t|s_t)) 
\end{align}

where the \textit{entropy} term is used to encourage exploration by penalizing agents' overconfidence. For the critic loss, the following expression is used:

\begin{align}
& \text{critic loss: } \frac{1}{2} \sum\limits_{t = 0}^{T - 1} {A(s_t, a_t)}^2 
\end{align}

\subsubsection{Deep Neural Networks}

As noted in \autoref{sec:relate_work}, the models and architectures that are currently used by the RL community to tackle combinatorial problems are well-suited for handling problems with variable state and action spaces. Therefore, this work relies on the Ptr-Net and the Transformer, which are robust and flexible architectures.
For brevity purposes, and due to space limitation, most of the mathematical foundations of these architectures are omitted and a quick overview is presented next.
For further details please refer to \cite{vinyals2015pointer} and \cite{vaswani2017attention}.

\paragraph{Pointer Network Architecture} \label{sec:ptr-net}

The main idea of Ptr-Net is that the output sequence is obtained by \textit{pointing} at elements in the input, hence the name of the architecture.
In other words, the output is a sequence of input selections, which makes Ptr-Net suitable for modeling selection problems.
For example, this architecture was used to solve classical problems like the TSP and knapsack problem \cite{vinyals2015pointer, gu2018pointer}.
In both cases, the solution is built by selecting the sequence by which cities are visited (in case of TSP) or sequence by which items will be placed into the backpack (in case of knapsack).

This architecture has two main components: the \textit{encoder} and the \textit{decoder}. The encoder's role is to map the input into a feature space, while the decoder generates the pointers.
In the encoding mode, the encoder reads the input data, one element at a time, and generates a set of fixed-dimensional vector representations, known as encoder hidden states $(e_1, ..., e_n)$ where $n$ is the number of input elements.
After reaching the end of the input sequence, Ptr-Net switches into a decoding mode.
In this mode, the decoder is initialized by the encoder's last hidden state $e_n$ and a user-defined start of sequence (SoS) symbol.
The decoder then produces the first hidden state $d_1$, which is used to generate the first pointer (i.e., index) to an element in the input sequence. 
In the next decoding step, $d_1$ and the previously pointed element are fed to the decoder  to generate a subsequent pointer.
This process repeats itself until the decoder points to a user-defined end of sequence (EoS) symbol or a predefined number of pointers is reached.
The decoder's hidden states are represented by $(d_1, ..., d_m)$ where $m$ is the size of the output sequence.

The actual pointing mechanism, which we call Ptr-Net head, is computed as follows.
First, at any decoding step $i$ the unnormalized log probabilities (logits) $u_i$ are calculated:

\begin{align} \label{eq:og-ptr-net}
u_i^j =  v^T \tanh(W_1 e_j + W_2 d_i), j \in (1, ..., n)
\end{align}

where $v^T, W_1, W_2$ are learnable parameters of the model. 

Next, $u_i$ is fed to a \texttt{softmax} layer to calculate the probabilities of pointers selection:

\begin{align} \label{eq:prob-ptr-net}
p_i =  \texttt{softmax}(u_i)
\end{align}

where $p_i$ is the softmax distribution with dictionary size equal to the length of the input.
Finally, a pointer is selected by sampling from $p_i$ via a sampling strategy (e.g., greedy, stochastic).
In greedy sampling, for example, the selected pointer is the index of the element with the highest probability in $p_i$.

It is worth to note that in the original Ptr-Net proposal \cite{vinyals2015pointer} both the encoder and the decoder use LSTMs.
However, the authors of Ptr-Net have shown later (same year) that RNNs, including LSTM and GRU, are not suited for solving problems where the order of the input does not matter \cite{vinyals2015order}.
% However, later on the same year, the authors of Ptr-Net have showed that RNNs, including LSTM and GRU, are not suited for solving problems where the order of the input does not matter\cite{vinyals2015order}.
The main limitation of RNNs is that they are order-sensible, an undesired behavior when working with sets of data.
Moreover, these networks are ineffective in learning long distance relationships between the elements in the input, \ie finding dependencies between elements that are located far apart in the input sequence\cite{hochreiter2001gradient}.
Since RNNs pass the information sequentially, the longer the input sequence, the more likely it is that some information will be lost during sequence processing.

\paragraph{Transformer Architecture} \label{sec:transformer}
This architecture, introduced in 2017, presents a new way of computing relationships between the elements in the input.
It completely abandons the use of RNNs and, instead, relies entirely on the attention mechanisms\cite{vaswani2017attention}.
Dropping the recurrence removed the constrains of sequential computation, allowing larger models to be built and a faster training to be achieved.
Current state-of-the-art NLP solutions (e.g., GPT-3\cite{brown2020language}) use the Transformer or one of its variants\cite{tay2020efficient}.

The core idea introduced by the Transformer is the self-attention mechanism that allows to find inter-dependencies between elements in the input, regardless of their positions in the input sequence.
Broadly speaking, self-attention receives as input an embedding representation of $n$ input elements, denoted by $(x_1, ..., x_n)$, and maps them into $(z_1, ..., z_n)$ output embeddings while preserving the input dimensions.
Each $z_i$ embedding, $i \in (1, ..., n)$, contains information about how input element at position $i$ is related to all the remaining input elements.
This input-output transformation process can be summarized as follows. 
First, every input element $x_i$, $i \in (1, ..., n)$, is mapped via linear transformation into key $k_i$, query $q_i$ and value $v_i$ vectors.
Next, a logit $u_i^j$ measuring the compatibility between $q_i$ and key $k_j$, $\forall j \in (1, ..., n)$, is computed:

\begin{align} \label{eq:og-transformer-logit}
u_i^j =  \frac{q_i k_j^T}{\sqrt{d^{(k)}}}, j \in (1, ..., n)
\end{align}
where $d^{(k)}$ is the dimension of $k$.

Then, just as in Ptr-Net, the compatibility vector $u_i$ is passed through a \texttt{softmax} layer (Expression (\ref{eq:prob-ptr-net})) to produce a probability vector $p_i$, whose size is equal to the length of the input.
Finally, the output $z_i$ is computed as:
\begin{align} \label{eq:og-transformer-softmax}
    z_i =  \sum_{j=1}^n p_i^j v_j, 
\end{align}
where $p_i^j$ is the $j$th weight in $p_i$.

The generic input-output mapping, provided by the self-attention mechanism, allow these computation blocks to be stacked to create a higher level of abstraction and, potentially, give the network more generalization capacity.
In the original work, authors stacked six self-attention blocks in the encoder and decoder and, at that time, were able to achieve state-of-the-art results in NLP tasks.

Please note that the Transformer was originally developed to solve NLP problems, where the order of the input sequence matters.
To that end, the authors introduced a so called \emph{positional encoding}\footnote{See "Section 3.5 - Positional Encoding" in \cite{vaswani2017attention}}.
Here in our work the order does not matter and, for this reason, the positional encoding is not included in our design.

\subsection{MDP Formulation} \label{sec:mdp_load_balancing}

% In this section, the RD problem in Definition \ref{def:RD_problem} will be formulated as an MDP, and described in terms of state, actions, and possible reward functions. 

In the following sections the RD problem, outlined in Definition \ref{def:RD_problem}, is formulated as an MDP and described in terms of state, actions, and possible reward functions. 

\subsubsection{State Space}

The state of the RD problem includes: $i$) a set of \texttt{Rules} that arrive at the reverse proxy and need to be distributed across nodes, denoted by $\mathcal{X}' \subseteq \mathcal{X}$; $ii$) a set of available nodes capable of hosting \texttt{Rules} in $\mathcal{X}'$, denoted by $\mathcal{N}' \subseteq \mathcal{N}$. 
Each \texttt{Rule} is represented by its profile that includes CPU, RAM and storage requirements, while each node is represented by its CPU, RAM and storage resources.
Note that the node set, i.e., locations where \texttt{Rules} can be placed, is also part of the action space, which will be explained in the following section.

For an adequate distribution of \texttt{Rules}, reverse proxy must have an up-to-date information about the number of available nodes and their states.
In distributed systems, nodes usually share this kind of information either by using passive or active methods.
In case of passive methods, nodes expose a specific endpoint, usually \texttt{/health}, that responds with its status: an \texttt{HTTP 200 OK} status code informs that the device is alive, which may be followed by information on available resources.
Then, the reverse proxy (e.g, Traefik) makes periodic requests to these endpoints and uses the obtained information during the distribution process.
In case of active methods, as in MoleculerJS \cite{moleculer}, nodes periodically broadcast \texttt{Heartbeat} packets containing information about their health status \cite{moleculer_discovery}.
Here we assume that a reverse proxy always has an up-to-date information on the available nodes and their CPU, RAM and storage resources.
Note that there might be situations where nodes do not have enough resources to serve all requests and, therefore, some of the requests will be rejected.
For this reason, we introduce a dummy node $node_\text{rej}$ where rejected \texttt{Rules} are placed.
This node does not have the ability to process the \texttt{Rules}, as it does not have any computational resources, and its purpose is to allow the RL agent to associate rejected \texttt{Rules} with a dedicated place.
The overall state for the RD problem is the following:

\begin{align}
\mathcal{S} \triangleq \{\mathcal{N}' \cup node_\text{rej},\mathcal{X}'\}.
\end{align}

Designing $\mathcal{S}$ this way allows the RL agent to have a global view of the problem, in order to make appropriate placement decisions.
Furthermore, this design can be applied to model other optimization problems (e.g, multiple knapsack problem, multiple vehicle routing problem) whose state is described by two sets and the problem solution can be modeled as a sequence of assignment decisions.

\subsubsection{Action Space}

It is assumed that a \texttt{Rule} $x \in \mathcal{X}'$ can be placed at any available node $n \in \mathcal{N}'$, as long as resources are not exceeded.
If a \texttt{Rule} has requirements higher than the resources available at the nodes in $\mathcal{N}'$ then it is rejected, i.e. placed at the $node_\text{rej}$.
Therefore, the action space is represented as follows:

\begin{align}
\mathcal{A} = \mathcal{N}' \cup node_\text{rej}
\end{align}

Placing a \texttt{Rule} at node $n$ involves updating the amount of available resources at $n$.
This is done by subtracting \texttt{Rule} resource requirements from node's available resources.
Placing a \texttt{Rule} at $node_\text{rej}$ does not involve any computation, as the \texttt{Rule} is being rejected.

From the user point-of-view, placing a \texttt{Rule} at any node in $\mathcal{N}'$ would result in a \texttt{HTTP 201 Created} meaning that the request was accepted and the \texttt{Rule} is running.
On the other hand, placing a \texttt{Rule} at the $node_\text{rej}$ would result in a \texttt{HTTP 503 Service Unavailable} response, meaning that the system is overloaded at the moment, and can not process the incoming request.
Therefore, the state and action spaces design is consistent with the realistic settings of \texttt{Rule} placement.

\subsubsection{Rewards}

The agent places \texttt{Rules}, one at a time, and in return receives a reward value.
This reward depends on the RD problem variant under consideration (see Section \ref{sec:math_load_balancing}).

\paragraph{Greedy Reward}

The reward signal for greedy optimization is simple and represented as follows:

\begin{align}
r(s_t, a_t) = \left\{\begin{matrix}
0, & \text{if \texttt{Rule} is rejected} \\ 
1, & \text{if \texttt{Rule} is accepted}
\end{matrix}\right.
\end{align}

Placing a \texttt{Rule} at any node in $\mathcal{N}'$ gives a positive reward to the agent, while placing it at $node_\text{rej}$, \ie rejecting the \texttt{Rule}, gives zero reward.
Since the agent aims to maximize the accumulated reward, it will try to come up with a distribution strategy that maximizes the number of placed \texttt{Rules}.

% If all the \texttt{Rules} in $\mathcal{X}'$ are placed at $\mathcal{N}'$ then the agent receives maximum

\paragraph{Critical-Aware Greedy Reward}

To maximize the placement of the \texttt{Rules}, while ensuring a fair distribution, the following reward is used:

\begin{align}
    r(s_t, a_t) = \left\{\begin{matrix}
    -2, & \text{if \texttt{Rule} is rejected} \\ 
    \Omega^{\text{MAX}}, & \text{if \texttt{Rule} is accepted}
    \end{matrix}\right.
    \label{equ:Critical-Aware-Greedy-Reward}
\end{align}

% For each \texttt{Rule} rejection the agent is penalized by receiving a negative two as a reward.
% On the other hand, placement of a \texttt{Rule} at node $n \in \mathcal{N}'$ gives reward equal to most critical resource among all the nodes present in $\mathcal{N}'$, as defined in Equation (\ref{eq:most_critical}). %$, at time step $t$.

% The way that state $S$ is modeled allows the agent to plan a placement strategy that, after distributing all \texttt{Rules}, maximizes the $\Omega^{\text{MAX}}$ and, therefore, maximizes the reward while ensuring that all the \texttt{Rules} are fairly distributed.

For each rejected \texttt{Rule} the agent is penalized by receiving a negative reward (-2, which has been selected empirically).
On the other hand, placing a \texttt{Rule} at node $n \in \mathcal{N}'$ gives a positive reward equal to the most critical resource among all nodes in $\mathcal{N}'$, as described in Expression (\ref{eq:most_critical}).

Penalizing rejections with negative rewards and rewarding successful \texttt{Rule} placement with positive rewards has two effects. %It encourages the placement of as much as possible \texttt{Rules} on the nodes and, at the same time, allows the agent to try to obtain high $\Omega^{\text{MAX}}$ rewards which brings fairness to the distribution of \texttt{Rules}.
It encourages the placement of the largest number of \texttt{Rules} and, at the same time, it allows the agent to obtain high $\Omega^{\text{MAX}}$ rewards, which brings fairness to the distribution of \texttt{Rules}.

\paragraph{Cost-Aware Greedy Reward}

The reward for maximizing the \texttt{Rule} placement while minimizing the number of used nodes is represented as follows:

\begin{align}
r(s_t, a_t) = \left\{\begin{matrix}
-2, & \text{if \texttt{Rule} is rejected} \\ 
-1, & \text{if \texttt{Rule} placed at an empty node} \\ 
 0, & \text{otherwise} \\ 
\end{matrix}\right.
\end{align}

The minimization of rejected requests is ensured by giving a negative reward, -2, for each rejected \texttt{Rule}.
To enforce that a minimum number of nodes is used, the agent receives a negative reward, -1, every time it places a \texttt{Rule} at an empty node, i.e., a node without previously assigned \texttt{Rules}.
Placing at a node that already has \texttt{Rules} assigned to it results in a reward equal to zero.
This way the agent will try to find a strategy that uses the lowest number of nodes to place all the \texttt{Rules}. 

\subsection{Model Architecture} \label{sec:rl_load_balancing}

\begin{figure*}\centering
\includegraphics[width=1.0\textwidth,trim={0.6cm 0.5cm 0.5cm 0.5cm}, clip]{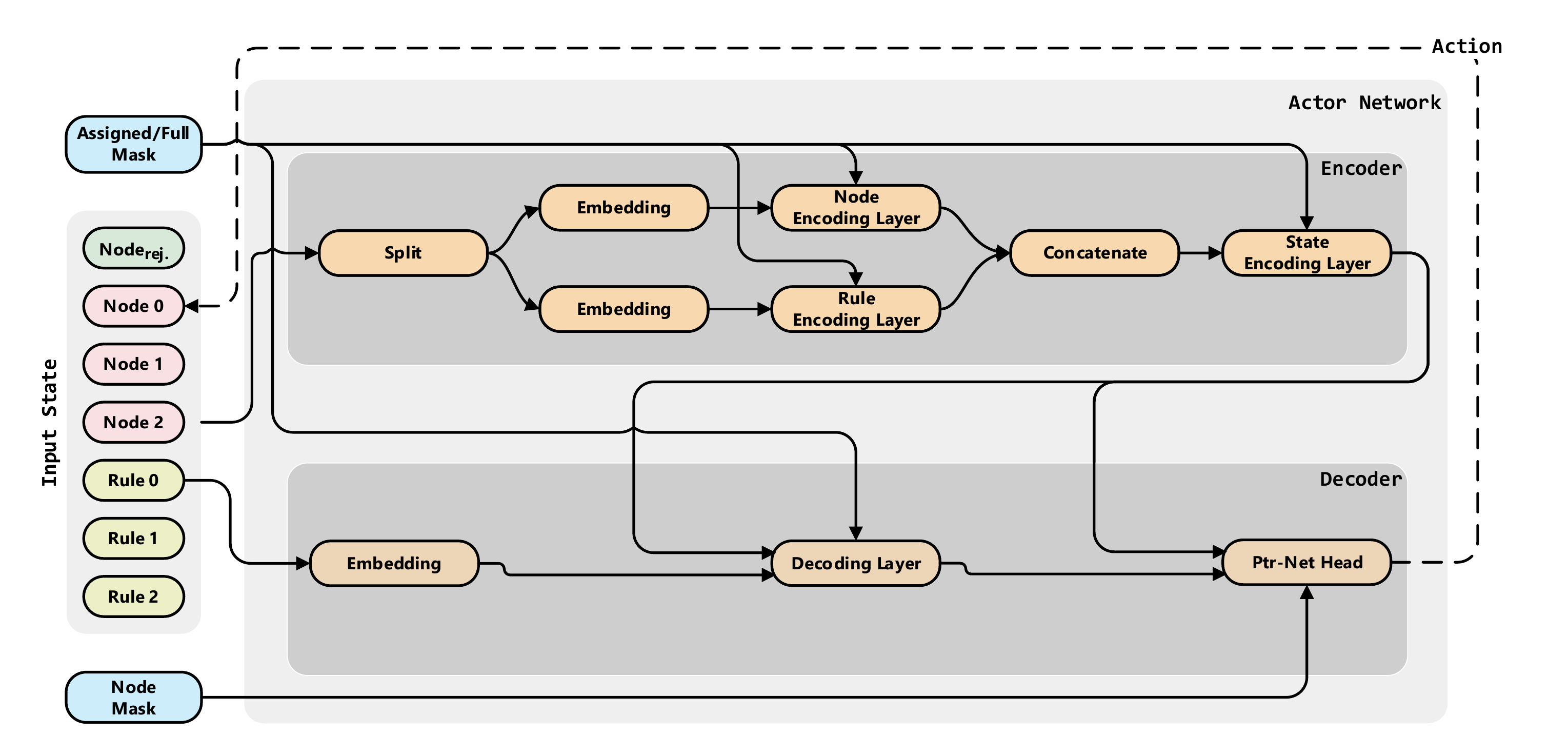}
\caption{Model architecture and decoding step example: \texttt{Rule 0} placed at node 0.}
\label{fig:modelArch}
\end{figure*}

\autoref{fig:modelArch} is a graphical representation of the proposed model, which has two main components (encoder and decoder) that are detailed in this section.
The source code of the model, implemented in Tensorflow\cite{tf2015}, is also publicly available\footnote{\url{https://github.com/AndreMaz/transformer-pointer-critic}}.

\subsubsection{Encoder}

The proposed encoder, illustrated in the upper part of \autoref{fig:modelArch}, uses self-attention mechanisms in all of its encoding layers.
%It receives $X = (x_1, x_2, ..., x_{n-m}, x_{n-m+1}, ..., x_n)$ elements as input, where $x_1$ is $node_\text{rej}$; $x_2$ to $x_{n-m}$ are the available nodes; $x_{n-m+1}$ to $x_n$ are the \texttt{Rules}.
It receives $\textbf{y} = (y_1, y_2, ..., y_{n-m}, y_{n-m+1}, ..., y_n)$ as input, where $y_1$ is $node_\text{rej}$; $y_2$ to $y_{n-m}$ are the available nodes; and $y_{n-m+1}$ to $y_n$ are the \texttt{Rules}.

In the proposed architecture, the input $\textbf{y}$ is split into two sets: \texttt{Rules} and nodes. Each set passes through a dedicated \texttt{embedding} layer and a self-attention block.
The decision of having two separate processing components, one for each set, has two reasons.
% First, to have a dedicated location in the network where it can look, in isolated manner, for relationships between the elements and find different strategies for pre-processing them.
% This pre-processing can be, for instance, sorting the rules and nodes using different strategies (as in traditional heuristic algorithms).
First, to give the model the ability to select different pre-processing strategies for each type of input.
A pre-processing can be, for instance, sorting the rules and nodes using different strategies (as in traditional heuristic algorithms).
However, in this case, the pre-processing strategies are not based on any human preconception, and the model learns adequate strategies that suit the QoS factor at hand.
The second reason is to make the model flexible to input data of uneven number of features. In other words, to make the model generic and applicable to different problems.
For example, for the cost-aware greedy optimization the node requires an additional binary feature in order to represent whether it is empty or not, meaning that the node is described by four features while the \texttt{Rule} is described by three only. 
Furthermore, the splitting idea can also be applied to more than two input components, which adds more flexibility to the proposed design. 

After completing the node and \texttt{Rule} pre-processing, the output is concatenated and passed into a final self-attention block where the network performs further processing. This time it looks for the relationships between the nodes and the \texttt{Rules}.

An interesting property of the proposed encoder is that it generalizes much better than the typical Transformer encoder, especially when the number of \texttt{Rules} or nodes increases.
This is mainly due to the flexibility of processing each input type separately using a dedicated location in the encoder.
% This is mainly due to the flexibility of our model (by splitting) to pre-process the rules and nodes in dedicated locations of the network.
Finally, to improve the training process and enforce problem constraints, the \texttt{Rules} already in place and full nodes are masked at each step.
Masking allows the encoder to discard specific elements during the processing stage.
The masking is done by modifying the logits calculation in Expression (\ref{eq:og-transformer-logit}) by: 

\begin{align} \label{eq:masked-transformer-logit}
u_i^j = \left\{\begin{matrix}
\frac{q_i k_j^T}{\sqrt{d^{(k)}}}, & j \in (1, ..., n-m), j \neq 1, j \neq \text{empty node}\\ 
-\infty, & \text{otherwise}
\end{matrix}\right.
\end{align}

The rule and the state encoding blocks use exactly the same expression but operate over different indices.
In the case of rule encoding block the \texttt{Rules} that are already assigned are masked.

\subsubsection{Decoder}
The proposed decoder is illustrated in the lower part of \autoref{fig:modelArch}.
At each step $i$, it receives encoder's output and a single \texttt{Rule} from range ($y_{n-m+1}, ..., y_n$) and generates a pointer to a node ($y_1, ..., y_{n-m}$).

In the decoding layer, the model looks for a (good) contender node for a given \texttt{Rule}.
Since the decoder has access to encoder's output, where the model had a global view of the problem, the selection of the contender takes also into consideration the presence of other \texttt{Rules} and the state of the nodes.
Finally, the last decoder layer is a Ptr-Net head, described in \autoref{sec:ptr-net}.
This head computes the location where the given \texttt{Rule} will be placed.
In order to avoid selecting unfeasible positions (e.g., full nodes), the logits computation of the Ptr-Net head in Expression (\ref{eq:og-ptr-net}) is modified as:

% OG expression
%\begin{align} \label{eq:ptn-head}
%u_i^j = \left\{\begin{matrix}
%v^T tanh(W_1 e_j + W_2 d_i), & j \in (1, ..., n-m, ..., m), j \leq n-m, j \neq 1, j \neq \texttt{Empty node}\\ 
%-\infty, & \text{otherwise} \\ 
%\end{matrix}\right.
%\end{align}

\begin{align} \label{eq:ptn-head}
u_i^j = \left\{\begin{matrix}
\textcolor{black}{v^T \tanh(W_1 e_j + W_2 d_i),} & j \in (1, ..., n-m, ..., n),\\ 
\textcolor{white}{v^T \tanh(W_1 e_j + W_2 d_i),} & j \leq n-m, j \neq 1,\\ 
\textcolor{white}{v^T \tanh(W_1 e_j + W_2 d_i)} & j \neq \text{empty node}\\ 
-\infty, & \text{otherwise} \\ 
\end{matrix}\right.
\end{align}

Furthermore, as in \cite{bello2016neural}, the logits from Expression (\ref{eq:ptn-head}) are clipped to $[-C_{\text{logit}}, C_{\text{logit}}]$, where $C_{\text{logit}} = 10$ in this work, to encourage the exploration:

\begin{align} 
u_i^\text{(clip)} = C_{\text{logit}} \cdot \tanh(u_i)
\end{align}

Then, the clipped logits $u_i^\text{(clip)}$ passes through a \texttt{softmax} layer, Expression (\ref{eq:prob-ptr-net}), to produce the probability vector from which the node is selected.
In the following decoding step, the resources of the selected node are updated, the assigned \texttt{Rule} is masked and the process repeats for the next \texttt{Rule} in $(y_{n-m+1}, ..., y_n)$.

It is important to outline that, in addition to sequential \texttt{Rule} feeding, we have tested random and neural-network based feeding strategies. 
However, other than adding additional complexity to the model we noticed no improvements in using any of these strategies. 
Since the model has a global view of the problem, via the encoder's output, the order by which the \texttt{Rules} are placed has no effect in the final result.

\section{Performance Evaluation} \label{sec:perfomance}

\subsection{Experimental Setup}

To evaluate the performance of the proposed agent, a comparison will be made against several  greedy heuristics and the optimal solution, generated using CPLEX solver. The optimal relies on the mathematical formalizations presented in \autoref{sec:math_load_balancing}.
The main goals are: $i)$ to assess the performance of the agent for different objective function criteria, and compare against baseline methods; $ii)$ to evaluate the scalability and generalization capacity of the agent, i.e., evaluate the performance of the agent when it has to solve problems larger than the ones used during training.
It worth to mention that performance evaluation does not consider other DRL approaches because, as mentioned in \autoref{sec:relate_work}, models used to solve task distribution and load balancing problems do not have the ability to scale, i.e., models have to be retrained each time the state or action space changes. Therefore, to validate the proposed method, optimal solutions are used as a baseline.

All the development and testing was done on a PC with i7 6700 3.4Ghz CPU with 32GB of RAM and a single Nvidia 2080Ti with 11GB GPU.

% The simulation environment was configured to generate: node instances on-the-fly, with CPU, RAM, and memory values sampled from a uniform distribution [0, 1]; a set of 1000 \texttt{Rules}, each with resource requirements sampled from uniform distribution between [0, 0.3]. The upper bound value for the \texttt{Rules} was selected to make the problem neither easy nor artificially hard, and to ensure adequate freedom during the \texttt{Rules} distribution process.
% Furthermore, to test the agent scalability, the environment was configured to produce problems of different sizes that are summarized below:

The simulation environment was configured to generate instances on-the-fly.
A uniform distribution, in range $[0, 1]$, is used to assign CPU, RAM, and storage resources to all nodes in an instance.
A pool of 1000 \texttt{Rules} is made available, for set sampling purposes, and resource requirement values are assigned using a uniform distribution in range $[0.01, 0.30]$.
This range ensures that the placement of \texttt{Rules} is neither too easy nor artificially hard, while providing some degree of freedom when distributing \texttt{Rules}.
Additionally, the environment was configured to produce problem instances of different sizes using the following parameters, allowing the assessment of the agent's performance and scalability:

\begin{itemize}
    \item \textit{Node-related parameters:}
        \begin{itemize}
        \item Smallest number of nodes: 10
        \item Largest number of nodes: 50
        \item Step size: 10
        \end{itemize}
    \item \textit{Rule-related parameters:}
        \begin{itemize}
        \item Smallest number of \texttt{Rules}: 10
        \item Largest number of \texttt{Rules}: 100
        \item Step size: 10
        \end{itemize}
\end{itemize}

The training process of the agent and baseline heuristics are discussed next, prior to the analysis of results.
It should be noted that the choice for relatively simple greedy heuristics, rather than complex heuristics or meta-heuristics, is due to the fact that it is not feasible to run complex approaches in real-time on reverse proxies.
As mentioned in \autoref{sec:proxy_balance}, the load balancing methods used in actual implementations are often simple strategies.
All strategies, RL agent and baseline heuristics, build a solution in a single pass, by starting from an empty solution and expanding it by placing one \texttt{Rule} at a time.

% Before discussing the results the agent training process and the heuristics, that were used as baselines, will be introduced.
% It is important to highlight that we did not consider comparing the agent against complex heuristics or meta-heuristics as it is not feasible to run them in real-time at reverse proxies.
% As mentioned earlier, current load balancing methods use simple strategies during the distribution process.
% Just as the proposed RL agent, all of the considered baseline heuristics construct a solution in a single pass, by starting from an empty solution and expanding it by placing a single \texttt{Rule} at a time.
% The goal is to investigate how RL-based load balancing strategies are comparable to heuristic-based strategies.

% In a pure qualitative comparison against specifically handcrafted heuristics, with no runtime considerations, methods such as beam-search or solution sampling could be used to improve the quality of the solution generated by the agent.
% Additionally, improvement heuristics (e.g., 2-opt) could be applied in order to refine even further the quality of agent's solution. However, such comparison is out of the scope of this work.

\subsection{Training and Hyperparameters}

The agent was trained on a fixed problem size of 10 nodes and 20 \texttt{Rules}, an empirically selected size that showed to be a good compromise between the training time, approximately 14 hours, and the agent's ability to generate robust solutions to all the problem instances during the testing.

The agent is trained using an A2C algorithm \cite{pmlr-v48-mniha16}, summarized in Algorithm \autoref{alg:a2c}, with mini-batches of size equal to 128, for better gradient estimates.
Two dedicated networks are used: $i$) the actor for learning the policy, denoted by $\pi_{\theta}$; $ii$) the critic to estimate the baseline, denoted by $V_{\phi}$.
The actor network is illustrated in \autoref{fig:modelArch} and described in detail in \autoref{sec:rl_load_balancing}.
The critic network is the encoder, from \autoref{fig:modelArch}, followed by three fully connected layers with linear activation.
The last layer of the critic network produces a single scalar value that represents the baseline estimation.
During the testing phase, only the actor network is used.

The choice for two separate networks, with no parameter sharing, was made in order to have more freedom to control agent's inference time and the quality of critic's baseline estimations.
The actor's encoder uses a single self-attention stack $N_{\text{actor}}^{\text{stack}}=1$ in each encoding block, while the critic uses a stack of three $N_{\text{critic}}^{\text{stack}}=3$.
Critic with less than three self-attention stacks produced poor baseline estimations.
Different learning rates were also used for the actor $\alpha_{\text{actor}} = 1e^{-4}$ and the critic $\alpha_{\text{critic}} = 5e^{-4}$ networks, parameters that have shown to be stable and to ensure convergence.
\autoref{tab:params} summarizes all the parameters that were adopted in this work.
Please note that the batch size and number of layer stacks in critic network were selected as a trade-off between the quality of the solution and the GPU memory constraints.

\begin{table}[htbp]
\small
  \centering
  \caption{Network(s) and Training Parameters.}
    \begin{tabular}{c|c}
    \hline
    \multicolumn{2}{c}{General} \\
    \hline
    %Parameter & Value \\
    %\hline
    \rowcolor{Gray}
    Training Steps & 100 000 \\
    Batch Size & 128 \\
    \rowcolor{Gray}
    Discount Factor ($\gamma$) & 0.99 \\
    Entropy Coefficient ($c_{\text{entropy}}$) & 0.01 \\
    \hline
    \hline
    \multicolumn{2}{c}{Actor Network} \\
    \hline
    %Parameter & Value \\
    %\hline
    \rowcolor{Gray}
    Weight Initialization & Xavier uniform\cite{glorot2010understanding} \\
    Number of Layer Stack ($N_{\text{actor}}^{\text{stack}}$) & 1 \\
    \rowcolor{Gray}
    Embedding Size & 128 \\
    Number of Heads & 8 \\
    \rowcolor{Gray}
    Inner Layer Dimension & 128 \\
    Logit Clipping ($C_{\text{logit}}$) & 10 \\
    \rowcolor{Gray}
    Gradient Clipping & L2 Norm (1.0) \\
    Optimizer & Adam \cite{kingma2014adam} \\
    \rowcolor{Gray}
    Learning Rate ($\alpha_{\text{actor}}$) & $1e^{-4}$ \\
    \hline
    \hline
    \multicolumn{2}{c}{Critic Network} \\
    \hline
    %Parameter & Value \\
    %\hline
    \rowcolor{Gray}
    Weight Initialization & Xavier uniform\cite{glorot2010understanding} \\
    Number of Layer Stacks ($N_{\text{critic}}^{\text{stack}}$) & 3 \\
    \rowcolor{Gray}
    Embedding Size & 128 \\
    Number of Heads & 8 \\
    \rowcolor{Gray}
    Inner Layer Dimension & 512 \\
    Gradient Clipping & L2 Norm (1.0) \\
    \rowcolor{Gray}
    Last Layers Dimensions & 128 \\
    Optimizer & Adam \cite{kingma2014adam} \\
    \rowcolor{Gray}
    Learning Rate ($\alpha_{\text{critic}}$) & $5e^{-4}$ \\
    \hline
    \hline
    \end{tabular}%
  \label{tab:params}%
\end{table}%

%As discussed in Section \ref{sec:RL_basics}, this is an actor-critic algorithm.
%In this work, in order to give the agent more freedom, two separated neural networks were used.
%One for the actor, denoted by $\pi_{\theta}$, and another network for the critic, denoted by $V_{\phi}$.
%The role of the actor network is to learn the agent policy, while the critic network role is to estimate the baseline (in order to stabilizes the training process).
%During the testing phase, the actor network is only used.

\begin{algorithm}[h]\label{alg:a2c}
\small

\caption{Actor-Critic training algorithm (adapted from \cite{bello2016neural}). }
\SetAlgoLined
\SetKwInOut{Input}{Input}\SetKwInOut{Output}{Output}

\Input{$T$=training steps, $B$=batch size}
\Output{Trained actor network $\pi_\theta$}

Initialize actor $\pi_\theta$ network parameters\;
Initialize critic $V_{\phi}$ network parameters\;

\For{$ t=1$ \KwTo $T$}{
    $s_i \leftarrow \texttt{SampleProblem()}, \forall i \in \left\{1, ..., B\right\}$
    
    $a_i, r_i \leftarrow \texttt{SolveProblem(} s_i, \pi_\theta \texttt{)}, \forall i \in \left\{1, ..., B\right\}$
    
    $b_i \leftarrow \texttt{EstimateBaseline(} s_i, V_{\phi} \texttt{)}, \forall i \in \left\{1, ..., B\right\}$

    $A \leftarrow \texttt{ComputeAdvantage(} r_i, b_i \texttt{)}$ // Expression (\ref{equ:advantage})
    
    $E \leftarrow \texttt{ComputeEntropy(} \pi_\theta \texttt{)}$ // Expression (\ref{eq:entropy}) 
    
    $g_\theta \leftarrow \frac{1}{B} \sum_{i=1}^B A \nabla_\theta \log(\pi_\theta (a_i|s_i) ) + c_{entropy} \times E$
    
    $L_\phi \leftarrow \frac{1}{B} \sum_{i=1}^B ||A ||^2_2$
    
    $\theta \leftarrow ADAM(\theta, g_\theta)$
    
    $\phi \leftarrow ADAM(\phi, \nabla_\phi L_\phi)$

    % $A \leftarrow \texttt{ComputeAdvantage(} r_i, b_i \texttt{)}$\tcp{Expression \ref{equ:advantage}}
    %aaa\tcp*{}
}
\end{algorithm}

\subsection{Baseline Heuristics}

\subsubsection{Random Insertion}
This heuristic randomly distributes the set of \texttt{Rules} across a set of available nodes.
More specifically, a \texttt{Rule} and a node are randomly picked from corresponding sets and then the feasibility of the pair is checked, i.e., a validation is performed to check if the resources available at the node are enough to satisfy the requirements of the \texttt{Rule}. 
If this condition is not met, another node is picked.
This process is repeated until all \texttt{Rules} are in place or marked as rejected.

\subsubsection{Critical Resource Insertion}
This is a two step heuristic, summarized in Algorithm \autoref{alg:critical_heuristic}, that first sorts the \texttt{Rules} by their resource requirements and then distributes them across the nodes, possibly giving priority to fairness ($\Omega^n$). The following four variations were generated:

\begin{itemize}
\item Descending Rules - Descending Critical (DR-DC) % Fair Placement (DR-Fair) % dominant resource ASC True node ASC True 
\item Descending Rules - Ascending Critical (DR-AC) % Node Number Placement (DR-Node) % dominant resource ASC True node ASC False
\item Ascending Rules - Descending Critical (AR-DC) % Fair Placement (AR-Fair) % dominant resource ASC False node ASC True
\item Ascending Rules - Ascending Critical (AR-AC) % Node Number Placement (AR-Node) % dominant resource ASC False node ASC False
\end{itemize}

%DR-DC and AR-DC ($C_{\text{order}}=`DESC'$) prioritize placement of \texttt{Rules} at locations that maximize the critical resource and therefore, prioritize fair distribution.
%DR-AC and AR-AC ($C_{\text{order}}=`ASC'$), on the other hand, prioritize placement of \texttt{Rules} that minimizes the $\Omega^n$, which potentially leads to the reduction of the number of the used nodes.
% The other variable, $X_{\text{order}}$, that generates additional heuristic variations simply control whether smaller or larger \texttt{Rules} are placed first.

DR-DC and AR-DC ($C_{\text{order}}$=`desc') prioritize the placement of \texttt{Rules} at locations that maximize the availability at the most critical resource, ensuring fairness. DR-AC and AR-AC ($C_{\text{order}}$ =`asc'), on the other hand, prioritize placement of \texttt{Rules} that minimizes the $\Omega^n$, which potentially leads to fewer nodes in use.
The  $X_{\text{order}}$ variable controls whether smaller or larger \texttt{Rules} are placed first.

\begin{algorithm}[h]\label{alg:critical_heuristic}
\small

\caption{Critical resource insertion heuristic.}
\SetAlgoLined
\SetKwInOut{Input}{Input}\SetKwInOut{Output}{Output}

\Input{$\mathcal{N}'$=node set, $\mathcal{X}'$=rule set, $X_{\text{order}} \in \text{\{`asc',`desc'\}}$ is the rule size sorting criteria, $C_{\text{order}} \in \text{\{`asc',`desc'\}}$ is the critical resource sorting criteria.}

\Output{Feasible solution}

$\mathcal{X}' = \texttt{SortByLargestResource(} \mathcal{X}', X_{\text{order}} \texttt{)}$

\ForEach{$x$ \text{in} $\mathcal{X}'$}{

    Initialize critical resource list $L_{\Omega}$

    \ForEach{$n$ \text{in} $\mathcal{N}'$}{
        // Determine critical resource
        
        $\Omega^n = \texttt{ComputeCritical(} n, x \texttt{)}$
        
        $L_{\Omega} \leftarrow (\Omega^n,n)$
    }
    
    // Sort $L_{\Omega}$ by $\Omega^n$
    
    $L_{\Omega} = \texttt{Sort(} L_{\Omega}, C_{\text{order}} \texttt{)}$
    
    // Do the first fit
    
    \ForEach{$\Omega^n$, $n$ \text{in} $L_{\Omega}$}{
    
        $allocated$ = False
        
        \uIf{$\Omega^n \geq 0$}{
            // Place \texttt{Rule} $x$ at node $n$
            
            $allocated$ = True
            
            $break$ // Break the loop
        }
    }
    
    \uIf{allocated == $False$}{
    
        Reject $x$
    }
}
\end{algorithm}

\subsection{Evaluation}

% In order to ensure that the agent's performance is consistent, and not a result of a lucky weight initialization, three different versions are obtained by training the agent with different seeds (1234, 1235 and 1236).
% After that, each agent solves 100 problem instances for each scenario under consideration.
% Therefore, the results reported here are an average of 300 executions.

% Moreover, to ensure that CPLEX results are obtained in feasible time, a time limit was set to 60 seconds.
% With this configuration CPLEX solver takes approximately 12 days to solve all the problem instances for a single QoS factor.

% Following the key performance indicators (KPIs) expressed mathematically and discussed in \autoref{opt:probs}, three main scenarios were considered.

Evaluation consists in solving three different RD variants, mathematically expressed in \autoref{opt:probs}, and observe how the agent performs in the key performance indicators (KPIs) of each.
To ensure the consistency of the agent performance, three different seeds were used to initialize the agent training weights, which essentially results in three distinct versions of the same agent.
After the training is finished, each agent solves 100 problem instances for each scenario under consideration.
Therefore, the reported performance is an average of the three agents, i.e., an average of 300 executions.
The baseline heuristics, which are deterministic methods, also solve 300 problem instances.
Finally, to ensure that results of the CPLEX solver are obtained within a feasible time, a time limit of 60 seconds is set. 
With time limit set, CPLEX solver will either return the optimal solution, if it manages to find one, or the best solution found so far.
For this configuration CPLEX takes approximately 8 days to solve all the problem instances of each RD variant under the consideration.

% Due to similarity and for brevity purposes, only results for node step

\subsubsection{Greedy Results} \label{results:greedy}

%The only KPI for the greedy optimizer (see \autoref{cplex:greedy}) is the \texttt{Rule} rejection rate, i.e., the likelihood of a user's request being rejected.
%\autoref{fig:greedy} shows the performance of the different solving methods for this particular case.
%When looking at the plots, the CPLEX solver presents the lowest rejection rate in all scenarios, as it provides the optimal solution, having the best performance for the KPI under consideration.

%%%%%%%%%%%%%%%%%%%%%%%%%%%%%%%%%%%%%%%%%%%%%%%%%%%%%%%%%%%%%%%%%%%%%%%%%%%%%%%%%%%%%%%%%%%%%%%%%%%%%%%%%%%%%%%%%%%%%%%%%%%%

\begin{figure*}\centering
\includegraphics[width=0.95\textwidth, clip]{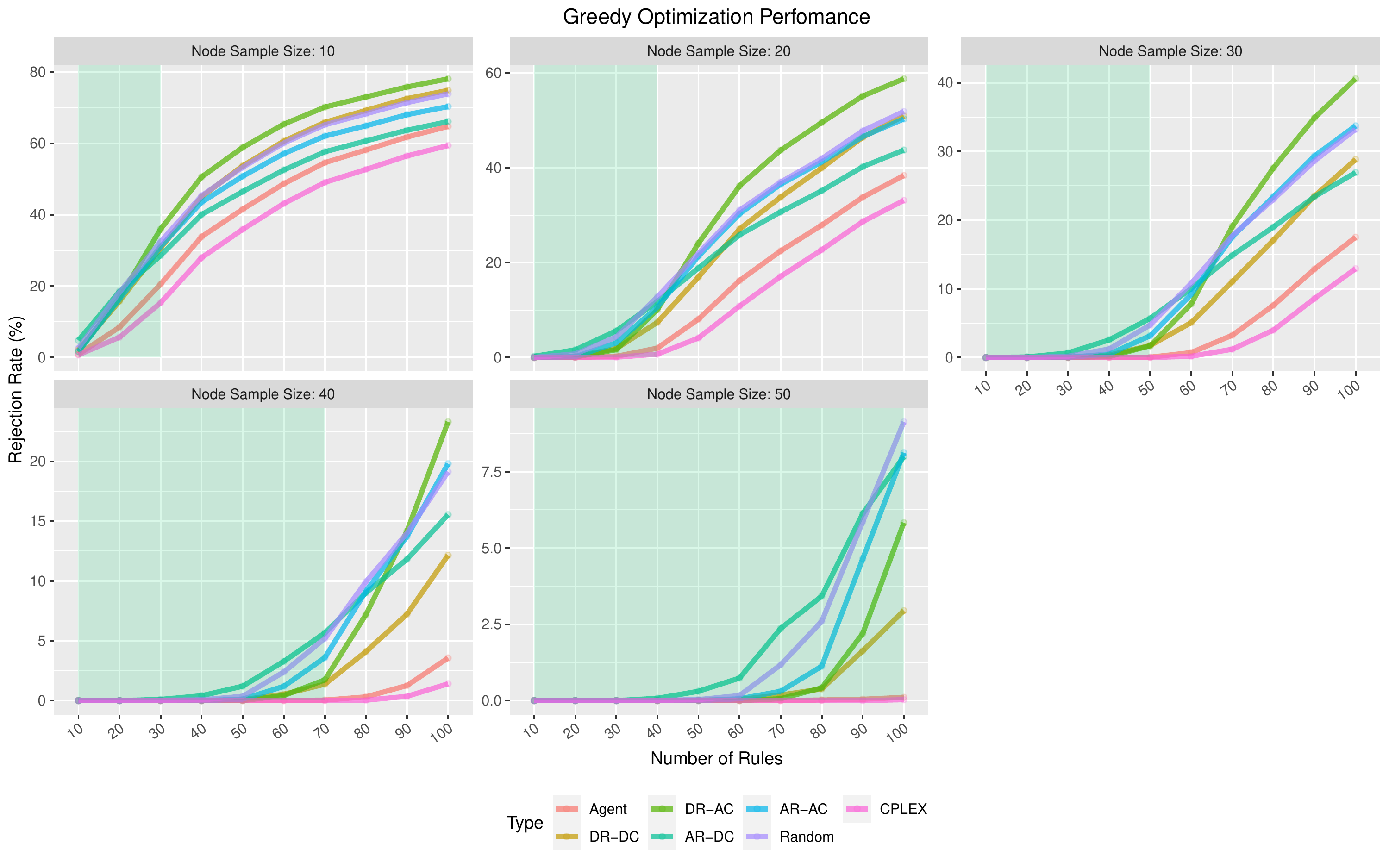}
% \caption{Evolution of \texttt{Rule} rejection rate for different problem sizes. Green highlighted areas outline the problem sizes where CPLEX was able to find optimal results in all test instances. In grey area CPLEX is considered an heuristic}
\caption{\texttt{Rule} rejection rate for different problem sizes. Green areas highlight where CPLEX was able to find optimal results for all instances, i.e., all the solution space was searched.}
\label{fig:greedy}
\end{figure*}

The KPI for the greedy optimizer (see \autoref{cplex:greedy}) is the \texttt{Rule} rejection rate, i.e., the likelihood of a user's request being rejected.
\autoref{fig:greedy} shows the performance of the different solving methods for this particular case.
Looking at the plots, it is visible that CPLEX solver has the lowest rejection rate in the all scenarios and, therefore, has the best performance.
Moreover, in the green highlighted areas CPLEX was able to find optimal solutions within the time limit.

% This can be explained by the fact that the mathematical formalization was especially handcrafted for this problem and because CPLEX solver had more time than any other method to explore the search space (more on execution time later).

The second best performing method is the proposed agent. 
%It shows similar performance to the CPLEX, especially in problem configurations where the latter was able to find optimal solutions.
It is visible that regardless of the problem size, the proposed agent has the smallest rejection gap with respect to the CPLEX solver, which acts as a lower bound.
As an example, for problems with 10 nodes the proposed agent has an average rejection rate 4.69\% higher than the CPLEX, as shown in \autoref{tab:greedy}.
As the number of available nodes increases the rejection gap becomes smaller, reaching an average of 0.01\% for problems with 50 nodes.
In contrast, the best performing heuristic (AR-DC) has an average rejection gap of 9.26\% for 10 nodes, when compared to CPLEX.
The DR-DC presents the lowest gap, 0.52\%, for 50 nodes, which is still larger than the agent's gap.

Results indicate that agent's generalization capacity is maintained even for large problems. 
Note that during training the agent solved problems with only 10 nodes and 20 \texttt{Rules}.
However, results do not show any critical degradation in performance, even for problem instances five times larger than the ones used in training.
For example, for the problems with 50 nodes, where CPLEX solves all problem instances to optimality, the agent generates near-optimal solutions.

\input{greedy_table.tex}

\subsubsection{Critical-Aware Greedy Results}

\begin{figure*}\centering
\includegraphics[width=1.0\textwidth, clip]{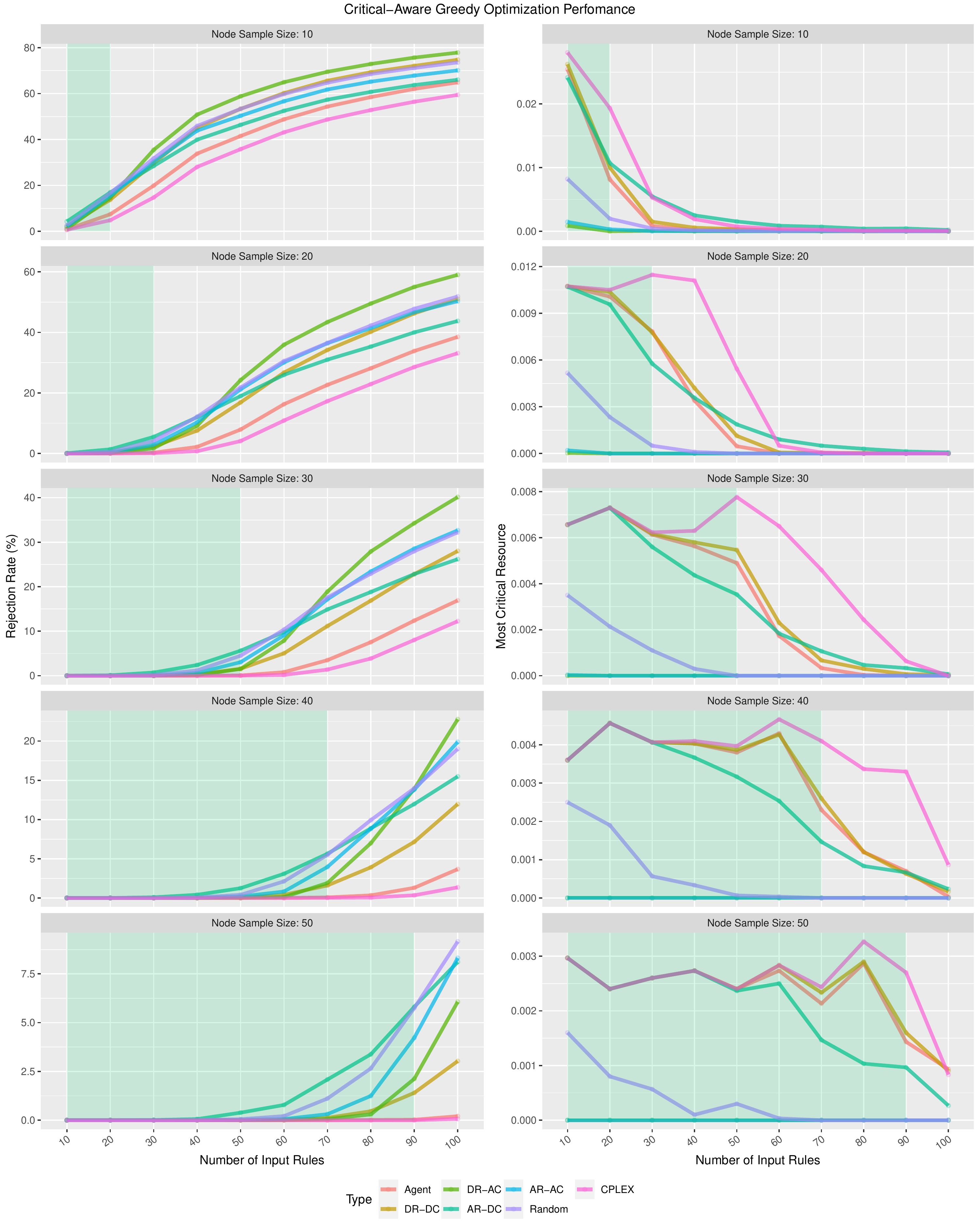}
%\caption{Evolution of \texttt{Rule} rejection rate (left column) and most critical resource (right column) for different problem sizes. Green highlighted areas outline the problem sizes where CPLEX was able to find optimal results in all test instances. In grey area CPLEX is considered an heuristic. Plots should by analyzed in horizontal pairs.}
\caption{\texttt{Rule} rejection rate (left column) and most critical resource (right column) for different problem sizes. Green areas highlight where CPLEX was able to find optimal results for all instances, i.e., all the solution space was searched.}
\label{fig:critical}
\end{figure*}

In this scenario, \autoref{cplex:crit_aware}, the \texttt{Rule} rejection rate and the most critical resource ($\Omega^{\text{MAX}}$) are the KPIs, which are shown in \autoref{fig:critical}.

As in the case of greedy optimization, the agent shows a similar trend when considering the rejection rate.
On the other hand, looking at the most critical resource at different ranges, the agent struggles to keep up with the CPLEX solver and seem to stay closer the DR-DC heuristic.
However, when taking both KPIs into account, it is visible that DR-DC's higher critical resource value is achieved at the expense of a higher rejection rate.
Similar behavior happens with the AR-DC heuristic.
Considering the KPI gaps between CPLEX and other methods, summarized in \autoref{tab:critical}, it is visible that in general the overall $\Omega^{\text{MAX}}$ differences are negligible but due to lower rejection rate the proposed agent offers a smoother user experience.

\input{critical_table.tex}

\subsubsection{Cost-Aware Greedy Results}

\begin{figure*}\centering
\includegraphics[width=1.0\textwidth, clip]{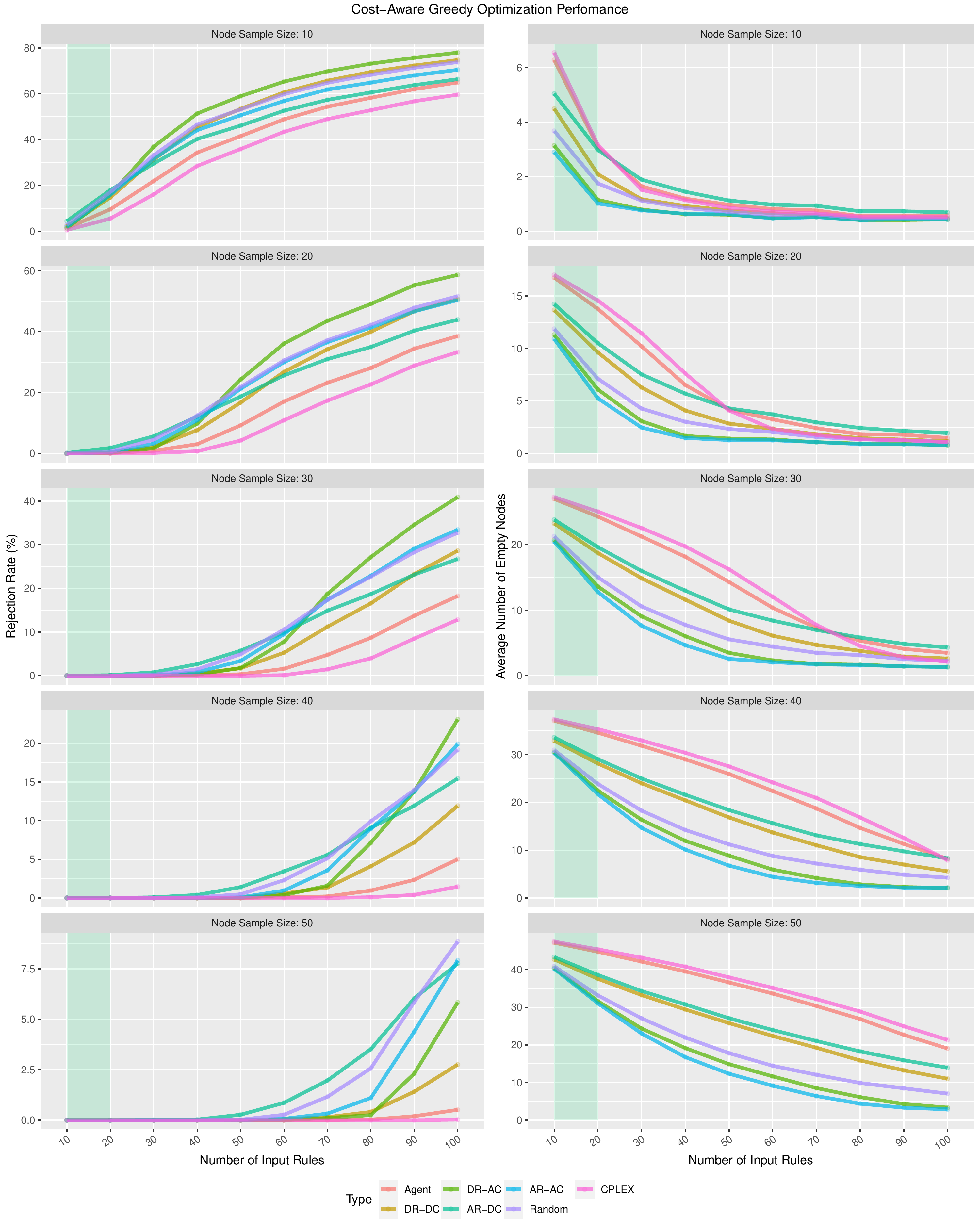}
% \caption{Evolution of \texttt{Rule} rejection (left column) and empty nodes (right column) for different problem sizes. Green highlighted areas outline the problem sizes where CPLEX was able to find optimal results in all test instances. In grey area CPLEX is considered an heuristic. Plots should by analyzed in horizontal pairs.}
\caption{\texttt{Rule} rejection rate (left column) and  empty nodes (right column) for different problem sizes. Green areas highlight where CPLEX was able to find optimal results for all instances, i.e., all the solution space was searched.}
\label{fig:cost}
\end{figure*}

For this RD optimization criteria, described in \autoref{cplex:node_aware}, the \texttt{Rule} rejection rate and the number of empty nodes are the KPIs, which are shown in the \autoref{fig:cost}.
% The left column represents the rejection rate while the right column the number of empty nodes in different problem configurations.

From all the RD problems under consideration, this one was the hardest for CPLEX.
This has to do with a greater number of binary variables involved in the mathematical formalization of the problem.
The solver did not manage to obtain optimal solutions for problems with more than 20 \texttt{Rules} to distribute.
% The apparent problem complexity might explain a larger rejection rate gap, summarized in \autoref{tab:cost}, between the agent and the CPLEX method when compared with the gap in critical-aware optimization shown in \autoref{fig:critical}.
The problem complexity might explain, therefore, the larger agent-CPLEX gap in the rejection rate, when compared with other optimization criteria \autoref{fig:critical}.
These results are summarized in \autoref{fig:critical}.
Nevertheless, the agent shows a trend similar to CPLEX solver in both KPIs and does not show critical performance degradation, as it happens with the heuristics.

When looking at gaps in \autoref{tab:cost}, it is possible to see that for 10 nodes the agent uses in average less -0.03 nodes but has a rejection rate is 4.9\% higher that the CPLEX.
As the number of nodes increases the rejection rate gap decreases at the expense of additional nodes in use.
For example, for 50 nodes the agent requires more 1.45 nodes to have almost the same rejection rate as the CPLEX.

In addition, when looking at the results of the heuristics, it is visible that they struggle to keep both KPIs gaps close to the CPLEX and the proposed agent.
In fact, from the three scenarios being considered, the cost-aware is where the heuristics struggle the most.
This highlights that a change of a single KPI requires redesigning and developing a new heuristic, which is impractical and time consuming.
In the case of the proposed agent, which performs well in all scenarios, the only change would be the reward signal.

\input{cost_table.tex}

\subsubsection{Inference Time} 
Reporting the time to obtain the solutions is important but hard to measure.
It depends on the hardware used (e.g., CPU vs GPU, GPU model), programming language, code implementation/optimization and so on. 
Nevertheless, a practical approach was taken and the reported times, summarized in \autoref{tab:times}, are averages of 1000 executions obtained on the development machine.
For brevity purposes, only the results for 100 \texttt{Rules} are shown, i.e., the upper bound.
Also, note that the reported measurements are rough approximations that can vary from execution to execution.
Still, they show the trend and the orders of magnitude that each method takes to compute the solution.
The running times of CPLEX are not reported because, as mentioned earlier, it is not feasible to run it at reverse proxies.
% Moreover, in many tested scenarios CPLEX execution was interrupted because of the 60 seconds time limit.
Finally, agent's reported times are the average duration of an entire episode, which also accounts the environment's transitions.

Looking at the agent's results it is visible that its time is practically constant, regardless of the number of nodes in the input.
This is a result of: $i)$ using the Transformer architecture that, as mentioned in \autoref{sec:transformer}, does not use RNNs to compute the relationships between the input elements; $ii)$ GPU parallelization capabilities, allowing to compute the relationships between all the elements in a single pass.
It is also worth to mention that, regardless of the tested input size, the model takes approximately 2.5 milliseconds in every decoding step, i.e., to find a location for a single \texttt{Rule}.

Comparing the agent's times with the heuristics, it is visible that they operate at different orders of magnitude.
For example, for 10 nodes all the heuristics are around 28 times faster that the agent.
However, for 50 nodes, this difference drops down to 6 for the critical resource heuristics and 20 for the random heuristic.
This time difference reduction is explained by the complexity of the heuristics: $O(r \cdot n\text{log}(n))$ for the critical resource heuristic and $O(r \cdot n)$ for the random heuristics, where $r$ is the number of \texttt{Rules} and the $n$ is the number of nodes.
Nevertheless, despite the reported time differences, the agent is able to generate high quality or even optimal solutions in 400 milliseconds, which makes the proposed model an appealing option.

Also, it should be noted that, although it is not visible in \autoref{tab:times}, the self-attention mechanism in the Transformer has $O(m^2)$ complexity, where $m$ is total number of input elements \cite{tay2020efficient}.
The reported constant time for the tested problem instances is ensured by the GPU parallelization.
Once the problem size become large enough, and hardware limitations of the GPU are reached, the effects of quadratic complexity become visible in the rapidly growing inference time.
However, this limitation can be tackled by replacing the Transformer with a recently introduced variation called the \textit{Linformer}\cite{wang2020linformer}, which reduces the self-attention complexity from $O(m^2)$ to $O(m)$ while maintaining similar performance in NLP tasks.
Overall, the performance and the inference speed of the proposed model can be improved even further by tuning the hyperparameters, using more advanced RL training algorithms (e.g., rolling baseline \cite{kool2018attention}, proximal policy \cite{schulman2017proximal}), using more recent variations of the Transformer \cite{tay2020efficient, tay2020long} and by doing model optimization\footnote{See \url{https://www.tensorflow.org/lite/performance/model_optimization}} and compression \cite{cheng2017survey}.

\begin{table}[]
\small
\centering
\caption{Time, in milliseconds, to obtain the solution for problems with 100 \texttt{Rules}. Reported times are an average of 1000 executions.}
\begin{tabular}{T|K|P|P|P|P|P}
\hline
Node Sample Size & Agent  & DC-DC & DC-AC & AR-DC & AR-AC & Random \\
\hline
\rowcolor{Gray}
10               & 394.82 & 14.49 & 14.14 & 14.22 & 14.35 & 12.27  \\
20               & 398.39 & 28.53 & 28.46 & 28.13 & 28.60 & 18.59  \\
\rowcolor{Gray}
30               & 402.89 & 42.28 & 42.61 & 41.46 & 42.45 & 20.05  \\
40               & 403.31 & 56.79 & 57.77 & 55.80 & 57.43 & 20.37  \\
\rowcolor{Gray}
50               & 404.17 & 69.86 & 72.24 & 69.42 & 71.80 & 21.14 
\end{tabular}
\label{tab:times}
\end{table}

\section{Conclusions} \label{sec:conclusions}
% In this work we investigate how RL-based load balancing strategies can be used by reverse proxy server to distribute event processing tasks among heterogeneous edge devices available for processing.

% This article investigated the use of DNN-based load balancing strategies, considering three different optimization criteria, for the distribution of event processing tasks across edge devices.

This article investigated the use of DNN-based strategies for the distribution of event processing tasks across edge devices, considering three different optimization criteria.
Each criteria is mathematically formulated, ensuring proper task execution, fair usage of resource and/or cost minimization related to device operation.
% Furthermore, MDP formulation to the aforementioned scenarios is introduced and followed by the presentation of the model, that is based on Transformer and Ptr-Net architectures.
Furthermore, an MDP formulation is introduced, which is followed by the presentation of the model based on Transformer and Ptr-Net architectures.

Simulation experiments show that the proposed model is capable of producing high quality solutions in all scenarios under consideration.
Additionally, it is scalable and capable of solving problems with up to five time larger than the ones that it was trained upon.
The latter is a very important ability as it allows to solve instances of different sizes without the need of retraining, which makes it easy to deploy in real world.
Existing work applying DNN and RL methods to load balancing tend to neglect model scalability and only work with static environment settings, which is not applicable in real world.

% Finally, the proposed model is generic and can be potentially applied to several problems that can be modeled as two sets of elements and solved by taking sequential assignments decisions.
In summary, the proposed model is generic and can be applied to problems that are modeled using two sets of elements and solved by taking sequential assignments decisions.
The latter is an important characteristic in load balancing problems as the increasing network complexity will likely generate several problem variations for which there is no good performing heuristic, making RL and machine learning methods attractive and of practical value.

Future research will focus on further tuning the performance of the proposed method while studying its applicability to other combinatorial problems that can be modeled in a similar fashion to the problem investigated in this work.

\section*{Acknowledgment}
This work was supported by FCT (Foundation for Science and Technology) from Portugal within CEOT (Center for Electronic, Optoelectronic and Telecommunications), by UID/MULTI/00631/2020 project and by FCT (Foundation for Science and Technology) Ph.D Grant SFRH/BD/138836/2018.



% \newpage 
% \bibliographystyle{IEEEtran}
\bibliographystyle{acm}
\bibliography{ref2}

\end{document}

%% file: greedy_table.tex
\begin{table}[htbp]
\small
\centering
\caption{Average rejection rate gap with respect to the CPLEX solver for the greedy optimization.}
\begin{tabular}{T|K|P|P|P|P|P}
%                 & \multicolumn{6}{c}{Rejection Gap}              \\
\hline
Node Sample Size & Agent & DR-DC & DR-AC & AR-DC & AR-AC & Random \\
\hline
\rowcolor{Gray}
10               & 4.69  & 14.4  & 17.97 & 9.26  & 11.95 & 14.44  \\
20               & 3.18  & 10.73 & 16.18 & 9.64  & 12.28 & 13.2   \\
\rowcolor{Gray}
30               & 1.51  & 6.08  & 10.49 & 7.64  & 9.01  & 9.27   \\
40               & 0.34  & 2.36  & 4.5   & 4.52  & 4.58  & 4.93   \\
\rowcolor{Gray}
50               & 0.01  & 0.52  & 0.85  & 2.1   & 1.42  & 1.89  
\end{tabular}
\label{tab:greedy}%
\end{table}

%% file: critical_table.tex
\begin{table*}[]
\small
\centering
\caption{Average $\Omega^{\text{MAX}}$ and rejection rate gaps with respect to the CPLEX solver for the critical-aware optimization.}
%\begin{tabular}{ccccccccccccc}
%\begin{tabular}{D|MM|MM|MM|MM|MM|MM}

\begin{tabular}{D|SS|SS|SS|SS|SS|SS}
\hline
                 & \multicolumn{2}{c}{Agent}             & \multicolumn{2}{c}{DC-DC}             & \multicolumn{2}{c}{DR-AC}             & \multicolumn{2}{c}{AR-DC}             & \multicolumn{2}{c}{AR-AC}             & \multicolumn{2}{c}{Random}            \\
\hline
Node Sample Size & $\Omega^{\text{MAX}}$ & Rej. Rate & $\Omega^{\text{MAX}}$ & Rej. Rate & $\Omega^{\text{MAX}}$ & Rej. Rate & $\Omega^{\text{MAX}}$ & Rej. Rate & $\Omega^{\text{MAX}}$ & Rej. Rate & $\Omega^{\text{MAX}}$ & Rej. Rate \\
\hline
\rowcolor{Gray}
10               & 0.00210               & 4.75          & 0.00175               & 14.14         & 0.00553               & 17.73         & 0.00093               & 9.18          & 0.00544               & 11.91         & 0.00453               & 14.38         \\
20               & 0.00173               & 3.21          & 0.00155               & 10.77         & 0.00498               & 16.04         & 0.00164               & 9.62          & 0.00496               & 12.18         & 0.00417               & 12.99         \\
\rowcolor{Gray}
30               & 0.00157               & 1.55          & 0.00137               & 6.01          & 0.00483               & 10.51         & 0.00172               & 7.57          & 0.00483               & 8.89          & 0.00413               & 9.12          \\
40               & 0.00080               & 0.36          & 0.00076               & 2.33          & 0.00366               & 4.39          & 0.00118               & 4.50          & 0.00366               & 4.57          & 0.00312               & 4.92          \\
\rowcolor{Gray}
50               & 0.00020               & 0.02          & 0.00015               & 0.50          & 0.00252               & 0.85          & 0.00059               & 2.06          & 0.00252               & 1.41          & 0.00218               & 1.89         
\end{tabular}
\label{tab:critical}%
\end{table*}

%% file: cost_table.tex
\begin{table*}[]
\small
\caption{Average empty node and rejection gaps with respect to the CPLEX solver for the cost-aware optimization.}
\centering
%\begin{tabular}{ccccccccccccc}
%\begin{tabular}{D|MM|MM|MM|MM|MM|MM}
\begin{tabular}{D|SS|SS|SS|SS|SS|SS}
\hline
                 & \multicolumn{2}{c}{Agent}       & \multicolumn{2}{c}{DC-DC}       & \multicolumn{2}{c}{DR-AC}       & \multicolumn{2}{c}{AR-DC}       & \multicolumn{2}{c}{AR-AC}       & \multicolumn{2}{c}{Random}      \\
\hline
Node Sample Size & Empty Nodes & Rej. Rate & Empty Nodes & Rej. Rate & Empty Nodes & Rej. Rate & Empty Nodes & Rej. Rate & Empty Nodes & Rej. Rate & Empty Nodes & Rej. Rate \\
\hline
\rowcolor{Gray}
10               & -0.03           & 4.9           & 0.41            & 14.19         & 0.76            & 17.84         & -0.04           & 9.09          & 0.8             & 11.82         & 0.55            & 14.28         \\
20               & 0.03            & 3.63          & 1.79            & 10.7          & 3.39            & 16.05         & 0.71            & 9.61          & 3.62            & 12.24         & 2.66            & 13.06         \\
\rowcolor{Gray}
30               & 0.45            & 2.06          & 4.31            & 6.04          & 7.86            & 10.43         & 2.72            & 7.58          & 8.38            & 8.97          & 6.41            & 9.14          \\
40               & 1.23            & 0.67          & 7.78            & 2.32          & 13.85           & 4.4           & 6.03            & 4.54          & 14.8            & 4.53          & 11.65           & 4.91          \\
\rowcolor{Gray}
50               & 1.45            & 0.07          & 10.67           & 0.48          & 19.28           & 0.85          & 9               & 2.05          & 20.75           & 1.38          & 16.44           & 1.87         
\end{tabular}
\label{tab:cost}%
\end{table*}